\documentclass[lettersize,journal]{IEEEtran}
\usepackage[ruled,vlined]{algorithm2e}
\usepackage{amsmath,amsfonts}
\usepackage{algorithmic}
\usepackage{array}
\usepackage[caption=false,font=normalsize,labelfont=sf,textfont=sf]{subfig}
\usepackage{textcomp}
\usepackage{stfloats}
\usepackage{url}
\usepackage{verbatim}
\usepackage{graphicx}
\usepackage{cite}

\begin{document}

\title{Bayesian optimization for sparse neural networks with trainable activation
functions}

\author{Mohamed Fakhfakh$^{1,2}$ and Lotfi Chaari$^{2}$ 

\begin{minipage}{8cm}
\centering
\vspace{0.2cm}
 \small $^1$ University of Sfax, MIRACL,  Tunisia\\
 \small mohamed.fakhfakh@toulouse-inp.fr
\end{minipage}
\begin{minipage}{8cm}
\centering
\vspace{0.2cm}
\small  $^2$ Toulouse INP, IRIT, University of Toulouse, France\\
 \small lotfi.chaari@toulouse-inp.fr
\end{minipage}
 }

\maketitle

\begin{abstract}
In the literature on deep neural networks, there is considerable interest in developing activation functions that can enhance neural network performance. In recent years, there has been renewed scientific interest in proposing activation functions that can be trained throughout the learning process, as they appear to improve network performance, especially by reducing overfitting. 
In this paper, we propose a trainable activation function whose parameters need to be estimated. A fully Bayesian model is developed to automatically estimate from the learning data both the model weights and activation function parameters. An MCMC-based optimization scheme is developed to build the inference. The proposed method aims to solve the aforementioned problems and improve convergence time by using an efficient sampling scheme that guarantees convergence to the global maximum. The proposed scheme is tested  on three datasets with three different CNNs. Promising results demonstrate the usefulness of our proposed approach in improving model accuracy due to the proposed activation function and Bayesian estimation of the parameters. 
\end{abstract}


\section{Introduction}
Classification is a machine-learning task that identifies which objects are present in an image or video. It is critical for various applications such as computer vision \cite{alsarhan2021machine,wang2020deep,dong2015image}, medical diagnostics \cite{sree2021novel}, signal processing \cite{jaini2021tool}, and others. The process involves learning significant or nontrivial relationships from a set of training data and extending these relationships to interpret new test data \cite{mitchell1996investigation}. The task involves categorizing elements into one of the finite set of classes by comparing the measured attributes of a given object with the known properties of objects to determine whether the object belongs to a specific category. \\
Convolutional Neural Networks (CNNs) \cite{drewek2021survey,sajja2021image,fakhfakh2020prognet,li2018deep,ji20123d} have become the industry standard in numerous applications over the past two decades, as they can process complex high-dimensional input data into simple low-dimensional concepts through a series of nonlinear transformations. Each feature layer in CNNs comprises features from the layer below, creating a hierarchical organization of ever-more-abstract concepts. They are particularly effective at capturing high-level abstractions in real-world observations, making them a popular choice for image classification tasks. CNNs can learn features from raw image pixels, reducing the need for manual feature engineering. This makes them well-suited for tasks such as object recognition, where the goal is to identify the presence and location of specific objects in an image. 
In this context, the activation function plays a critical role in learning representative features. The Rectified Linear Unit (ReLU) \cite{hara2015analysis} is currently the most widely used activation function for neural networks. When provided with a positive argument, the ReLU activation function is either zero or the identity. ReLUs have the additional benefit of alleviating the vanishing gradient problem, in addition to providing sparse codes \cite{lau2018review}. According to the state of the art, activation functions can be fixed or trainable during a learning phase \cite{apicella2021survey}. In most cases, gradient descent is the most widely used method in the literature for parameter estimation. \\
From another side, Bayesian methods have advanced significantly in many domains over time and have numerous useful applications. The primary idea is to represent all uncertainties in the model using probabilities. Bayesian techniques are distinctive in that they treat the problem as an inference problem \cite{butts2003network}. One of the most significant advantages is the ability to incorporate prior information about the model parameters and hyperparameters. Recent advancements in Markov Chain Monte Carlo (MCMC) methods \cite{andrieu2010particle,robert2013monte,chaari14,chaari19} make it easier to use Bayesian analyses in complex datasets with missing observations and to handle multidimensional outcomes. Recent studies have demonstrated that using a Bayesian framework in CNNs for the optimization process leads to more promising performances than standard gradient descent \cite{fakhfakh2022efficient,fakhfakh2022bayesian}. \\
In this study, we introduce a new trainable activation function. The parameters of the proposed function are automatically estimated from the data. For doing so, a Bayesian framework is used where these parameters as well as the network weights are assumed to be realizations of random variables. With an adequate likelihood, a hierarchical Bayesian model is built with priors and hyperpriors. An MCMC-based inference is then use, specifically a Gibbs sampler, to derive estimators from the target distributions. 
Our method is an extension of our previous work described in \cite{fakhfakh2022nonsmooth}, where we employed non-smooth Hamiltonian methods to fit sparse artificial neural networks. Our main objective with the Bayesian scheme is to minimize the target cost function of the learning model. The use of non-smooth Hamiltonian techniques enables us to perform efficient and fast sampling, even when dealing with non-differentiable energy functions that arise due to the use of sparse regularization functions. \\
The contribution of this paper is therefore twofold: \emph{i)} proposing a new trainable activation function, and \emph{ii)} a more general Bayesian formulation than in \cite{fakhfakh2022nonsmooth} to integrate the estimation of all parameters from the data.\\
The rest of this paper is organized as follows. The state of the art is the focus of Section \ref{sec:related_work}. Then, the Problem statement is in Section \ref{sec:problem}. In section \ref{sec:hierarchical} we detail the adopted hierarchical Bayesian model. The proposed Bayesian inference scheme is developed in Section \ref{sec:inference} and validated in Section \ref{sec:validation}. 
Finally, The conclusion and future work are drawn in Section \ref{sec:conlusion}.

\section{Related work}
\label{sec:related_work}
Finding the best activation function to integrate into an architecture is a challenging task. This section proposes a taxonomy of different activation functions described in the literature. 
A primary established classification is based on the ability to modify the shape of the activation function during the training phase, resulting in two major categories that can be distinguished \cite{apicella2021survey}.
\subsection{Fixed-shape activation functions}
This category pertains to the use of activation functions in neural network research, including sigmoid \cite{university1988continuous}, hyperbolic tangent (tanh) \cite{xiao2005simple}, and ReLU, all of which have a defined form. The introduction of rectified functions, particularly ReLU, has led to a marked enhancement in neural network performance and heightened scientific interest. Consequently, this category can be subdivided into subcategories based on their distinctive characteristics.\\

\textbf{Classic activation functions:} \\
The findings presented in \cite{cybenko1989approximation} demonstrated that a feed-forward and shallow network can efficiently manage any continuous function defined on a compact subset. For many years, bounded activation functions like Sigmoid and tanh were the preferred choices for neural networks, with researchers demonstrating their efficacy, particularly in shallow network architectures \cite{ding2018activation}. Although Sigmoid, Bipolar sigmoid \cite{harrington1993sigmoid}, Hyperbolic tangent, Absolute value \cite{chadha2002fractional}, and other bounded activation functions are commonly used, their efficacy is limited when training multi-layer neural networks due to the vanishing gradient problem \cite{bengio1994learning}. \\

\textbf{Rectifier-based activation functions:} \\
The primary advantage of using rectified activation functions is to alleviate the problem of vanishing gradient. The success of ReLU \cite{glorot2011deep} has inspired the development of many new activation functions during the last years \cite{gulcehre2016noisy,nie2011multistability}. 
While ReLU has numerous benefits such as solving the vanishing gradient issue and making sparse coding easier \cite{montalto2015linear}, it is not without flaws. The "dying" ReLU problem \cite{maas2013rectifier} and non-differentiability at zero are the main concerns.\\
To address these issues, several variations of ReLU have been developed, such as Leaky ReLU (LReLU) \cite{maas2013rectifier}, Truncated rectified \cite{konda2014zero}, softplus  \cite{dugas2000incorporating}, Exponential linear unit (ELU) \cite{Shah2016deep}, E-swish \cite{ccelebi2020new}, and Flatten-T Swish \cite{chieng2018flatten}.

\subsection{Trainable activation functions} 
The concept of using trainable activation functions is not new in the field of neural network research, with many studies published on this topic as early as the 1990s \cite{chen1996feedforward, guarnieri1995multilayer, piazza1992artificial}. However, the growing interest in neural networks in recent years has led researchers to reconsider the potential benefits of trainable activation functions in improving network performance.\\
In this section, we discuss the main strategies proposed in the literature for learning activation functions from data. Based on their primary characteristics, these strategies can be classified into three families. \\

\textbf{Parameterized standard activation functions:} \\
In \cite{chen1996feedforward}, a generalized hyperbolic tangent function was proposed by introducing two trainable parameters $\alpha$ and $\beta$, which adjust the saturation level and slope, respectively. Similarly, a sigmoid function with two trainable parameters was used in \cite{guarnieri1995multilayer} to modify the activation function's shape. These parameters are learned along with the network weights using the backpropagation algorithm. More recently, the work in \cite{trottier2017parametric} aimed to avoid manually setting the parameter of the ELU unit by proposing an alternative based on two trainable parameters. 
The proposed activation function, called PELU, is defined as follows:

\begin{equation}
PELU(x) = 
\begin{cases}
\frac{\beta}{\gamma} x                & \text{if $ x \geq 0 $}  \\
\beta \times (exp(\frac{x}{\gamma}) - 1)   &\text{otherwise,} 
\end{cases}
\label{eq:e38}
\end{equation}

where $\beta$ is a trainable parameter.\\

A flexible ReLU function has been proposed in \cite{qiu2018fre}:

\begin{equation}
frelu(x) = ReLU(x + \alpha) + \beta ,
\label{eq:e39}
\end{equation}

where $\alpha$ and $\beta$ are parameters learned from data. This is done to capture negative information that is lost in the classic ReLU function \cite{hara2015analysis}. 


The activation function introduced in \cite{he2015delving} is another type of ReLU function that partially learns its shape from the training set. Indeed, it can modify the negative part of the data via the parameter $\alpha$. This function is called Parametric ReLU (PReLU) and can be defined as follows:

\begin{equation}
PReLU(x) =
\begin{cases}
x & \text{if $ x > 0 $} \\
\alpha \times x &\text{otherwise}
\end{cases}
\end{equation}

where the parameter $\alpha$ is learned jointly with the  model using a gradient method. \\

\textbf{Functions based on ensemble methods:} \\
Functions based on ensemble methods involve combining multiple basic activation functions to form a more complex function. In \cite{nader2020searching}, a method for investigating activation functions built as compositions of various basic activation functions are proposed. Similarly, \cite{basirat2018quest} introduces a similar method using a genetic algorithm composed of two new activation functions: Exponential Linear Sigmoid SquasHing (EliSH) and HardELiSH. EliSH is defined as follows:


\begin{equation}
EliSH(x) =
\begin{cases}
x / (1 + e^{-x}) & \text{if $x \geq 0$} \\
(e^x - 1) / (1 + e^{-x}) & \text{otherwise}.
\end{cases}
\end{equation}

The negative part of EliSH is a multiplication of two functions, ELU and Sigmoid, while the positive part is shared with Swish. HardELiSH is defined as a multiplication of HardSigmoid and ELU in the negative portion and HardSigmoid and Linear in the positive part. In \cite{maguolo2021ense}, another interesting activation function is proposed, called the Mexican Hat Linear Unit (MeLU). This activation function solves the problems of unstable learning related to trainable parameters. Unstable learning can cause a decrease in accuracy and an increase in generalization error when the model's performance varies significantly in response to slight changes in the data or parameters. Mexican hat-type functions have a smoother curve than ReLU, which prevents saturation and allows for optimal performance. \\
Let $f$  be the function defined by 

\begin{equation}
f_{\gamma,\lambda}(x) = max(\lambda - \vert x-\gamma \vert, 0),
\label{eq:e313}
\end{equation}

where $\lambda$, $\gamma$ are real numbers. This function returns zero when $\lvert x - \gamma \vert > \lambda$. Moreover, it increases with a derivative of 1 between $\gamma - \lambda$ and $\gamma$, then decreases using a derivative of -1 between $\gamma$ and $\gamma + \lambda$. MeLU is defined for each layer as

\begin{equation}
MeLU(x) = PReLU(x) + \sum_{j=1}^{k-1} c_{j} f_{\gamma_{j},\lambda_{j}}(x),
\end{equation}

where $k$ represents the number of parameters that can be learned in each neuron. 
The $c_j$ parameters are learnable real numbers, while  $\gamma_j$, $\lambda_j$ are fixed parameters. \\
In \cite{sutfeld2020adapt}, the authors introduced a new function called Adaptive Blending Units (ABU) as a trainable linear combination of a set of activation functions 

\begin{equation}
ABU(x) = \sum_{i=1}^{k} \alpha_i \times f_i(x),
\end{equation}

where ($\alpha_1$, $\alpha_2$,..., $\alpha_k$) are parameters to be learned, and \{$f_1(.)$, $f_1(.)$, ..., $f_k(.)$\} is a set of activation functions. 
The parameters $\alpha_i$ are all initialized to the value $\frac{1}{k}$ and are trained using the gradient descent method. \\
Likewise, similar methods were introduced recently like Kernel-based activation function \cite{scardapane2019kafnets,scardapane2018complex} and Trained activation function \cite{ertuugrul2018novel}. \\

\textbf{Activation functions based on other techniques:} \\
In \cite{piazza1992artificial}, a method using polynomial functions with adjustable coefficients is proposed. Similarly, in the context of fuzzy activation functions, a neural unit based on Type-2 fuzzy logic \cite{castillo2014review} was developed. 
Other works have proposed functions using interpolation and spline approaches \cite{scardapane2017learning}. However, depending on the chosen technique, these strategies may require additional input. 
\subsection{Comparison and analysis}
Learning activation functions is a popular topic in the field of machine learning because the performance of learning architectures can be improved by using more suitable activation functions. Various approaches have been explored to enhance these performances. Most trainable activation functions are variants of standard activation functions, whose shape is adjusted using trainable parameters. Several studies on trainable activation functions have reported substantial performance improvements compared to neural network architectures equipped with classic fixed activation functions such as ReLU or sigmoid. \\
Other trainable activation functions can be expressed as sub-networks nested within the main architectures or those based on different approaches than classical activation functions, such as fuzzy logic. 

While these activation functions have significant potential to improve neural network model performance, their implementation can be more complex and require more time and resources for learning.\\
Despite encouraging results, it is still challenging to identify a strategy for automatic learning of an activation function that would solve different problems and significantly improve performance. Most trainable activation functions use the gradient descent method for hyperparameter estimation. The main limitations of these techniques lie in the computation time and gradient vanishing. This process prevents the network from learning deep features and may even lead to excessive processing capacity during training.\\
Indeed, neural networks can get stuck in local minima \cite{wang2019reltanh}, which can harm the model's performance. This phenomenon is partly due to the gradient vanishing that occurs as derivatives decrease as the model deepens. This gradient decrease makes it harder to optimize the model, leading to a decrease in performance.

\section{Problem statement}
\label{sec:problem}
In the previous Section, we have examined different activation functions presented in the literature. 
In this paper, we introduce a modification of the MeLU activation function \cite{maguolo2021ensemble}, while integrating the estimation of its parameters into a global Bayesian optimization framework. The choice of this function is mainly justified by its promising performance, as well as its form which promotes non-linearity and sparsity. However, the limits of the MeLU function are mainly related to memory requirements. This limit indicates that using this function may require a larger amount of memory compared to other activation functions, which can affect computational efficiency. \\
As a first contribution of this paper, the Modified Mexican ReLU (MMeLU) activation function is proposed to solve the complexity problem and improve model performance. The specificity of this new activation function is that it requires fewer parameters to estimate than MeLU. The second contribution is related to integrating the parameters estimation of the MMeLU function into a Bayesian optimizer \cite{fakhfakh2022bayesian1}, 
rather than using a standard optimization procedure such as the ADAM optimizer. \\
To define MMeLU, let

\begin{equation}
f_{\gamma,b}(x) = max(b - \vert x - \gamma \vert, 0), 
\end{equation}

where $\gamma$ and $b$ are real numbers. \\
Using $f_{\gamma,b}$, the proposed activation function MMeLU can be defined as

\begin{equation}
\textit{MMeLU}(x) = c \times f_{\gamma, b}(x) + (1 - c)  \times \textit{ReLU}(x)
\end{equation}

where $c$ is a real number belonging to the interval $[0,1]$. 
For the proposed MMeLU function, $c$, $b$, and $\gamma$ are the parameters to be estimated.\\

\begin{figure}[!htp]

\begin{center}
\begin{tabular}{c c}
\hspace{-0.5cm}\includegraphics[height=4cm,width=4.5cm ]{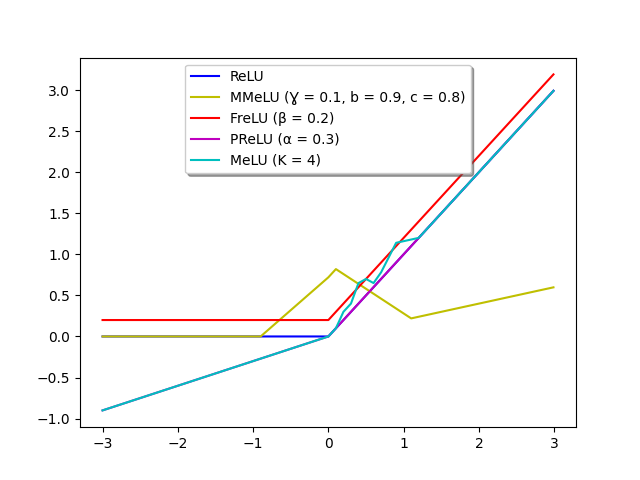} &
\hspace{-0.5cm}\includegraphics[height=4cm,width=4.5cm ]{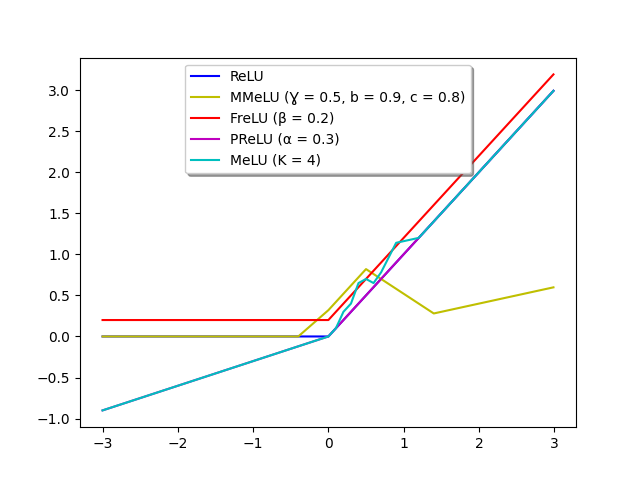} \\
(a) & (b) \\
\hspace{-0.5cm}\includegraphics[height=4cm,width=4.5cm ]{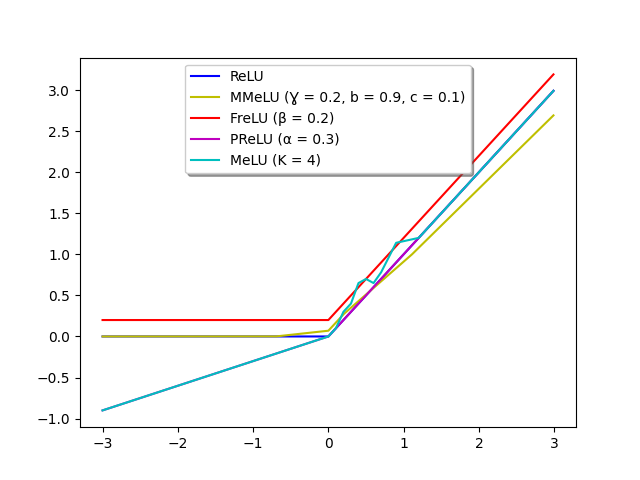} &
\hspace{-0.5cm}\includegraphics[height=4cm,width=4.5cm ]{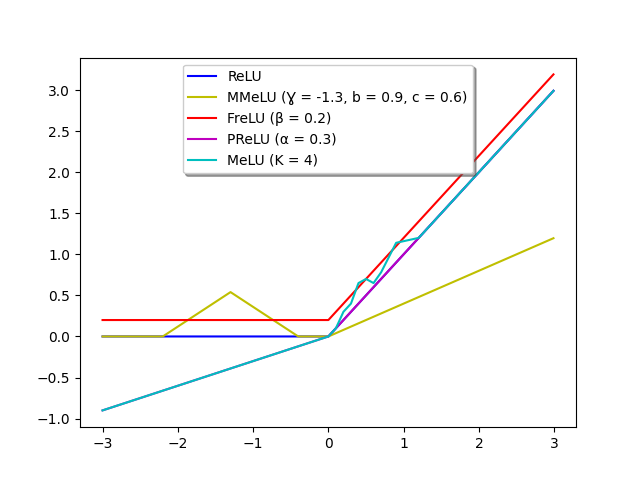} \\
(c) & (d) \\

\end{tabular}
\caption{\centering Illustrations of ReLU, FreLU, PReLU, MeLU, and MMeLU curves with different configurations. \label{fig:illust}} 
\end{center}
\end{figure}

The shape of the proposed MMeLU function, as well as those of competing functions ReLU, FReLU, PReLU, and MeLU, are illustrated in Figure \ref{fig:illust}. The activation function proposed in this paper is made up of a mixture of the ReLU and the Mexican hat functions. The curves in Figure \ref{fig:illust}[(a)-(d)] clearly show the flexibility and non-linearity of our MMeLU function with different configurations of the parameters $\gamma$, $b$, and $c$. \\
The Mexican hat function is often used as an activation function in neural networks due to its advantages. Mexican hat functions are continuous, which means that changes in inputs produce continuous changes in outputs. This is important in neural networks because it allows  a gradual update of weights. Additionally, they have high representation capacity, which means that they can accurately model complex functions. This allows neural networks to learn non-linear relationships between inputs and outputs.\\
The Mexican hat function looks like a bell but with a peak in the center that makes it more pronounced for small values, as shown by the MMeLU curves in Figures \ref{fig:illust}[a] and [b]. This means that for small input values, the Mexican hat function will have a stronger response than other activation functions, such as ReLU or FReLU. This behavior can be useful in certain situations, for example when the input data has a restricted value range or when the neural network's response needs to be more sensitive to small variations in the input data. \\

\begin{figure*}[!htp]
\begin{center}
\begin{tabular}{c}
\includegraphics[height=6.5cm,width=16cm ]{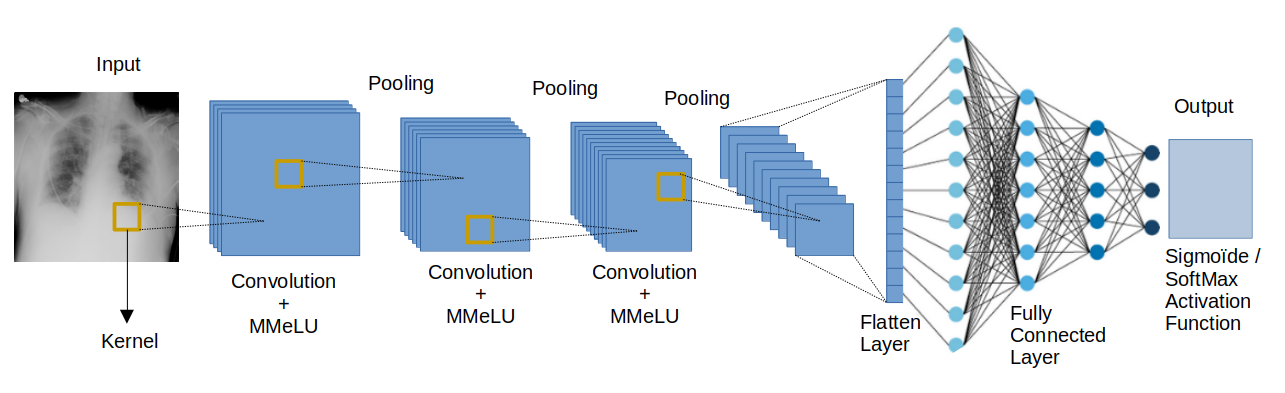}
\end{tabular}
\caption{\centering A general diagram of a convolutional neural network with the MMeLU activation function. \label{fig:mmelu}} 
\end{center}
\end{figure*}

Let us now consider the convolutional neural network in Figure \ref{fig:mmelu}. The MMeLU activation function is applied after each convolutional layer, replacing the \textit{ReLU} function. This strategy has the potential to increase the non-linearity of the model and improve its ability to represent features in images. It is worth noting that when $c$ is estimated close to 0 ($c\sim 0$), the MMeLU function tends to the ReLU behavior. \\

Regarding the model fitting, let us assume that the  estimated label (or numerical value) is obtained by applying the proposed activation function MMeLU($\textbf{x}$,$W$), where $\textbf{x}$ is the input data and $W \in \mathbb{R}^N$ denotes the weights vector. The model parameters (weight vector and parameters of the activation function) can be determined during the training phase using a generic error function $D$ (Euclidean, Minkowski, etc.). Akin to \cite{fakhfakh2022bayesian}, the target CNN is assumed to be sparse. We also use the same Bayesian formulation of the optimization problem. For the $M$ input data, we can write

\begin{align}
\label{eq:reg}
&\widehat{W} = \arg \min_{W} \mathcal{L}(W) \nonumber \\ 
&= \arg \min_{W,b,\gamma,c}  \sum_{m=1}^{M} D( MMeLU(x^m;W) - y^{m} ) + \sum_{l=1}^{L} \lambda_l \Vert W^l \Vert_1, 
\end{align} 

where $y^m$ is the ground truth for input data $x^m$, $L$ is the number of layers in the network, and $\lambda_l$ is a regularization parameter to be estimated for layer $l$ that balances the solution between the data attachment term and the $\ell_1$ sparse regularization terms.\\
In the following Section, we formulate the adopted hierarchical Bayesian model to conduct the inference and fit the model weights and the activation function parameters. 

\section{Hierarchical Bayesian Model}
\label{sec:hierarchical}
The problem of estimating the parameters of the MMeLU activation function is formulated within a Bayesian framework. In this sense, all parameters and hyperparameters are supposed to follow probability distributions. A likelihood is defined to model the relationship between the target weight vector, the activation function parameters, and the data. A prior distribution is defined to model the prior knowledge about the target weights and all activation function parameters.

\subsection{Likelihood}
Following the principle of minimizing the error between the reference vector $\textbf{y}$ (labels or continuous values) and its estimate $\widehat{\textbf{y}}$, we define the likelihood distribution as

\begin{align}
&f(\textbf{y},\textbf{x};W,c,\gamma,b) \propto \nonumber \\
& \prod_{m=1}^{M} \exp \left[- D ; ( MMeLU(x^m;W,c,\gamma,b) - y^{m} ) \right].
\end{align}
It is worth noting that when a Euclidean distance is used for $D$, the adopted likelihood is nothing but a Gaussian distribution. 
\subsection{Priors}
In our model, the unknown parameters are grouped in the unknown vector $\theta$ = \{$W$, $c$, $\gamma$, $b$, $\lambda$\}, where $\lambda = \{\lambda_1,\ldots,\lambda_L\}$. \\

\textbf{Prior for $W$:} \\
To promote sparsity in the neural network, we use a Laplace distribution for the weight vector $W$ akin to \cite{fakhfakh2022bayesian}:

\begin{equation}
f(W;\lambda) \propto \prod_{l=1}^{L} \prod_{k=1}^{K_l} \left[\frac{1}{\lambda_l} \exp \left( - \frac{\rvert W_k^l \rvert}{\lambda_l} \right) \right],
\end{equation}

where $K_l$ is the number of weights in layer $l$ of the network and $\lambda_l$ is a parameter to be estimated.\\

\textbf{Prior for $\lambda_l$:} \\
Since $\lambda_l \in \mathbb{R}_+$, we chose to use an inverse gamma (IG) distribution:

\begin{align}
f(\lambda_l;\delta,\mu) = IG(\lambda_l;\delta,\mu) \propto (\lambda_l)^{-1-\delta} \exp \left(-\dfrac{\mu}{\lambda_l} \right),
\end{align}

where $\delta$ and $\mu$ are positive parameters that were fixed at $10^{-3}$ to have a non-informative prior. \\

\textbf{Prior for $c$:} \\
Regarding the parameter $c$, we consider a uniform distribution over the interval $[0, 1 ]$, denoted as

\begin{equation}
c \sim U_{[0, 1]} (c).
\end{equation}

\textbf{Prior for $\gamma$ :} \\
Since $\gamma$ is a real value, a Gaussian distribution is used as follows \\

\begin{equation}
f(\gamma;\sigma^2) = \frac{1}{\sqrt{2 \pi \sigma^2}} \exp \left(-\frac{\gamma^2}{2 \sigma^2} \right),
\end{equation}

where $\sigma^2$ is a hyperparameter to be estimated.\\

\textbf{Priori for $b$:} \\
Since $b$ is a positive real number, an exponential distribution is used as follows:

\begin{equation}
f(b;\lambda_b) \propto
\begin{cases}
\frac{1}{\lambda_b} \exp \left(- \dfrac{ b }{\lambda_b}\right) ; \text{if} ; b \geq 0 \\
0 ; \text{otherwise}.
\end{cases}
\end{equation}

where $\lambda_b$ is a hyperparameter to be estimated. This prior penalizes large values of $b$.

\subsection{Hyperpriors}
Since $\lambda_b$ and $\sigma^2$ are positive real numbers, an inverse gamma (IG) distribution was used as a hyper-\emph{a priori}: \\


\begin{align}
f(\lambda_b;\delta,\mu) = IG(\lambda_b;\delta,\mu) \propto (\lambda_b)^{-1-\delta} \exp \left(-\dfrac{\mu}{\lambda_b} \right)
\end{align}
and 

\begin{align}
f(\sigma^2;\delta,\mu) = IG(\sigma^2;\delta,\mu) \propto (\sigma^2)^{-1-\delta} \exp \left(-\dfrac{\mu}{\sigma^2} \right),
\end{align}

where $\delta$ and $\mu$ are positive parameters that were fixed at $10^{-3}$. \\

\section{Inference scheme}
\label{sec:inference}
By adopting a Maximum \emph{a Posteriori} (MAP) approach, we first need to express the posterior distribution. Let $\Phi_{e}$ be the hyperparameters to be estimated, represented by $\Phi_{e}$ = \{$\sigma^2$, $\lambda_b$\}, and $\Phi_{m}$ be the hyperparameters to be fixed, $\Phi_{m}$ = \{$\delta$, $\mu$\}. Using the likelihood, the prior distributions, and the defined hyperpriors, we can write the posterior distribution as:

\begin{align}
&f(\theta,\Phi_{e};y,\Phi_{m})\propto f(y;\theta)f(\theta;\Phi_{e})f(\Phi_{e};\Phi_{m}) \nonumber \\
\end{align}

which can be reformulated in a detailed version as

\begin{align}
&f(\theta,\Phi_{e};\textbf{y},\textbf{x},\Phi_{m})\propto \nonumber \\ &\prod_{m=1}^{M} \exp \left[- D \; ( MMeLU(x^m;W,c,\gamma,b)  - y^{m} ) \right] \times \nonumber \\
&\prod_{l=1}^{L} \frac{1}{\lambda_l^{K_l}} \prod_{k=1}^{K_l} \left[ \exp \left(- \frac{\rvert W_k^l \rvert}{\lambda_l} \right) \right] \times (\lambda_l)^{-1-\delta} \exp \left(-\dfrac{\mu}{\lambda_l} \right) \times\nonumber \\ 
&  \exp\left(-\frac{\gamma^2}{2 \sigma^2}\right) \times \frac{1}{\lambda_b} \exp \left(- \dfrac{ b }{\lambda_b}\right) {1}_{\mathbb{R}_+}(b) \times {1}_{[0,1]}(c) \times \nonumber \\ 
&  (\lambda_b)^{-1-\delta} \exp \left(-\dfrac{\mu}{\lambda_b} \right) \times (\sigma^2)^{-1-\delta} \exp \left(-\dfrac{\mu}{\sigma^2} \right). \nonumber \\
\label{eq:posterior}
\end{align}

One can clearly notice that the posterior in \eqref{eq:posterior} is complicated to deal with in order to derive close-form estimators. We, therefore, resort to numerical approximations using a Markov Chain Monte Carlo technique (MCMC) \cite{robert2013monte,Chaari_tsp_2016}. Specifically, we use a Gibbs sampler to sequentially sample according to the conditional posteriors. 
To calculate the conditional distributions associated with each parameter of the model, one needs to integrate the joint posterior distribution in \eqref{eq:posterior} with respect to all the other parameters.

Regarding the parameter $W$, calculations based on \eqref{eq:posterior} lead to the following form:
\begin{align}
&f(W;c,\gamma,b,\lambda) \propto \exp \left[- \sum_{l=1}^L \sum_{k=1}^{K_l} \frac{\rvert W_k^l \rvert}{\lambda_l} \right] \times \nonumber \\ 
&\exp \left[- \sum_{m=1}^M \left(D ; (MMeLU(x^m;W,c,\gamma,b) - y^{m}) \right) \right].
\end{align}

The conditional distribution for the parameter $c$ is given by:
\begin{align}
&f(c;W,b,\gamma) \propto {1}_{[0,1]}(c) \times \nonumber \\ 
&\exp \left[- \sum_{m=1}^M \left(D ; (MMeLU(x^m;W,c,\gamma,b) - y^{m}) \right) \right].
\end{align}

For the parameter $b$, the condition distribution is given by:
\begin{align}
&f(b;W,c,\gamma,\lambda_b) \propto \exp \left(- \frac{b}{\lambda_b} \right) \times \nonumber \\
&\exp \left[- \sum_{m=1}^M \left(D ; (MMeLU(x^m;W,c,\gamma,b) - y^{m}) \right)\right].
\end{align}

As regards $\gamma$, the conditional distribution writes:
\begin{align}
&f(\gamma;W,b,c,\sigma^2) \propto \exp \left(- \frac{\gamma^2}{2 \sigma^2} \right) \times \nonumber \\ 
&\exp \left[- \sum_{m=1}^M \left(D  (MMeLU(x^m;W,c,\gamma,b) - y^{m}) \right)\right].
\end{align}

The conditional distribution for the parameter $\lambda_l$ is given by:

\begin{align}
f(\lambda_l;\delta,\mu) &\propto  \lambda_l^{-1-(\delta+K_l)} \exp \left(-\dfrac{\mu}{\lambda_l} \right)  \nonumber \\
& \propto IG(\delta + K_l,\mu).
\end{align}
For the hyperparameter vector $\Phi_{e}$, it is necessary to calculate the conditional distributions from which it is possible to sample based on the likelihood and adopted priors.

The conditional distribution for the hyperparameter $\lambda_b$ is given by:
\begin{align}
f(\lambda_b;,b,\mu,\delta) &\propto \lambda_b^{-2-\delta} \exp \left(- \frac{b+\mu}{\lambda_b} \right) \nonumber \\
& \propto IG(\delta + 1,b+\mu)
\end{align}

The conditional distribution for the hyperparameter $\sigma^2$ is given by:
\begin{align}
f(\sigma^2;\mu,\gamma,\delta) &\propto (\sigma^2)^{-1-\delta} \exp \left(- \frac{\gamma^2+2\mu}{2 \sigma^2} \right) \nonumber \\
& \propto IG(\delta ,\gamma +2\mu).
\end{align}

The sampling scheme is summarized in Algorithm \ref{algo:Melu}, where the model weights $W$ and the parameters of the proposed MMeLU function are sampled. 



\begin{algorithm}
\SetAlgoLined
Fix the hyperparameters $\Phi_{m}$ ;\\
\For{$r=1,\ldots, S$}{
* Sample $c$ according to $f(c;W,b,\gamma)$ \;
* Sample $\gamma$ according to $f(\gamma;W,b,c,\sigma^2)$ \;
* Sample $b$ according to $f(b;W,c,\gamma,\lambda_b)$ \;
* Sample $\sigma^2$ according to $f(\sigma^2;\mu,\gamma,\delta)$ \;
* Sample $\lambda_b$ according to $f(\lambda_b;,b,\mu,\delta)$ \;
* Sample $\lambda_l$ according to $f(\lambda_l;\delta,\mu)$  $\forall \; l \in \{1,\ldots,L\}$  \;
* Sample $W$ as described in \cite{fakhfakh2022nonsmooth} \;
}
\caption{Gibbs sampler for the proposed method.} \label{algo:Melu}
\end{algorithm}

In Algorithm \ref{algo:Melu}, $S$ denotes the number of MCMC sampling iterations. After the burn-in period, the sampled coefficients are used to calculate the estimators $\widehat{W}$, $\widehat{c}$, $\widehat{b}$, $\widehat{\gamma}$, in addition to $\widehat{\sigma^2}$, $\widehat{\lambda}$ and $\widehat{\lambda_b}$.

\section{Experimental validation}
\label{sec:validation}
In order to validate the proposed method, three image classification experiments are conducted using different  datasets: COVID-19 dataset including Computed tomography (CT) images for challenging classification \cite{afshar2021covid}, and two standard datasets, namely Fashion-MNIST \cite{kayed2020classification} and CIFAR-10 \cite{ccalik2018cifar}. 
Table \ref{tab:datasets} illustrates the setting details of the different datasets.

\begin{table}[!htp]
\begin{center}
\caption{\footnotesize \label{tab:datasets} Setting details of the used datasets.}
\begin{tabular}{cccc} 
\hline
\vspace{0.2cm}
\textbf{Dataset} & \textbf{Training set} &   \textbf{Test set}  &  \textbf{\# Classes}   \\ 
\hline
CT images classification & 566 & 180  & 2 \\
\hline
Fashion-MNIST & 48000 & 12000 & 10 \\
\hline
CIFAR-10 & 50000 & 10000 & 10 \\
\hline
\end{tabular}
\end{center}
\end{table}

To compare the proposed method with the state of the art, two types of optimizers are used: \emph{i)} our previous Bayesian optimizer that uses non-smooth Hamiltonian methods described in \cite{fakhfakh2022nonsmooth} with the standard ReLU activation function, and \emph{ii)} the standard Adam optimizer (with a learning rate of $10^{-3}$) with eight of the most well-known activation functions: ReLU, LReLU, ELU, PReLU, SeLU \cite{klambauer2017self}, swish, FReLU, and MeLU.
As regards coding, we used python programming language with Keras and Tensorflow libraries on an Intel(R) Core(TM) i7-2720QM CPU 2.20GHZ architecture with 16 Go memory.

\subsection{ConvNet models}
In this work, three CNN architectures are utilized. Similar to the LeNet model \cite{lecun1998gradient}, the first one has two fully-connected and three convolutional (Conv-32, Conv-64, and Conv-128) layers (FC-64 and FC-softmax). The second architecture has nine convolutional (3XConv-32, 3XConv-64, and 3XConv-128) and three FC layers (FC-128, FC-64 and FC-softmax) which are organized similarly to VGG-Net \cite{muhammad2018pre}. These architectures are shown in Table \ref{tab:convnet1}. The third model is a deeper CNN with 25 convolutional layers and 4 FC layers (for more information, see section \ref{sec:deep_cnns}). Each one uses convolutional layers with a stride size of 1 and $3 \times 3$ filters in addition to $2 \times 2$ max-pooling. \\
Deep neural networks are expanded with three regularizing strategies since they can easily overfit when trained on small datasets: Batch Normalization \cite{ioffe2015batch}, $\ell_{1}$ Regularization \cite{xu20101} and Dropout \cite{srivastava2014dropout}. \\
We chose to use different architecture depths in the experiments, mainly to test the ability of our proposed method to achieve better performance in the shallow model. In this sense, training with large and complex data can be expensive.

\begin{table}[!htp]
\begin{center}
\caption{\footnotesize \label{tab:convnet1} Convnet with regularization techniques.}
\begin{tabular}{ c c } 
\hline
\vspace{0.2cm}
\textbf{CNN\_1} \quad \quad & \quad \quad  \textbf{CNN\_2} \\ 
Conv3x3-32:stride=1 \quad \quad & \quad \quad  3 X Conv3x3-32:stride=1 \\ 
BatchNormalization  \quad \quad & \quad \quad  BatchNormalization \\ 
MaxPool 2x2  \quad \quad & \quad \quad  MaxPool 2x2 \\ 
Dropout(0.2) \quad \quad & \quad \quad  Dropout(0.3) \\ 
                  
\vspace{0.1cm}
                  & \quad \quad   \\

Conv3x3-64:stride=1 \quad \quad & \quad \quad 3 X Conv3x3-64:stride=1 \\ 
BatchNormalization \quad \quad & \quad \quad  BatchNormalization \\ 
MaxPool 2x2 \quad \quad & \quad \quad  MaxPool 2x2 \\ 
Dropout(0.3) \quad \quad & \quad \quad  Dropout(0.3) \\ 

\vspace{0.1cm}
                  & \quad \quad   \\
                   
Conv3x3-128:stride=1 \quad \quad & \quad \quad 3 X Conv3x3-128:stride=1 \\ 
BatchNormalization \quad \quad & \quad \quad  BatchNormalization \\ 
MaxPool 2x2 \quad \quad & \quad \quad  MaxPool 2x2 \\ 
Dropout(0.4) \quad \quad & \quad \quad  Dropout(0.4) \\ 

\vspace{0.1cm}
                  & \quad \quad   \\                   
\vspace{0.1cm}
Flattening \quad \quad & \quad \quad Flattening\\
                   
FC-64 \quad \quad & \quad \quad FC-128 \\
Dropout(0.3) \quad \quad & \quad \quad Dropout(0.35) \\
                  & \quad \quad FC-64 \\
\vspace{0.1cm}
                  & \quad \quad Dropout(0.35) \\
                
FC-softmax \quad \quad & \quad \quad FC-softmax \\
\hline
\end{tabular}
\end{center}
\end{table}

\subsection{Sampling Results}
\begin{figure}[!htp]
\begin{center}
\begin{tabular}{c c}
\hspace{-0.5cm}\includegraphics[height=4cm,width=4.5cm ]{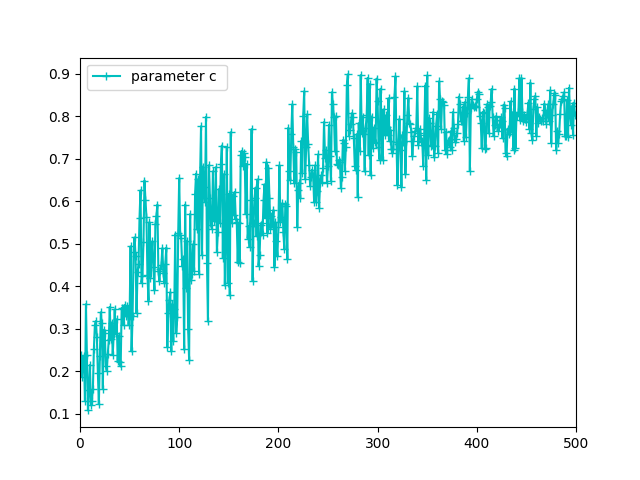} &
\hspace{-0.5cm}\includegraphics[height=4cm,width=4.5cm ]{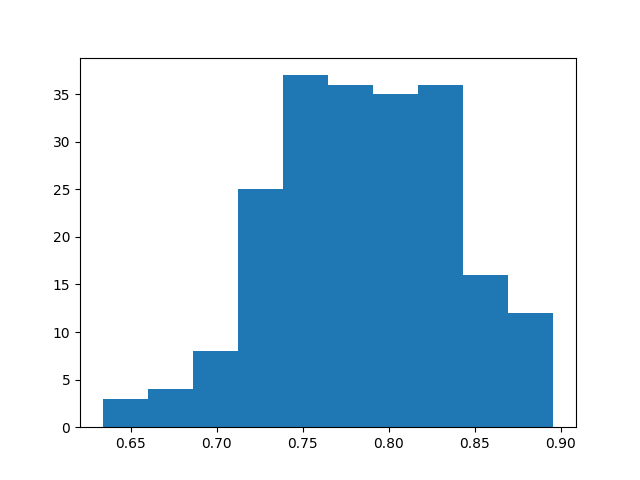} \\
(a) & (b) \\

\hspace{-0.5cm}\includegraphics[height=4cm,width=4.5cm ]{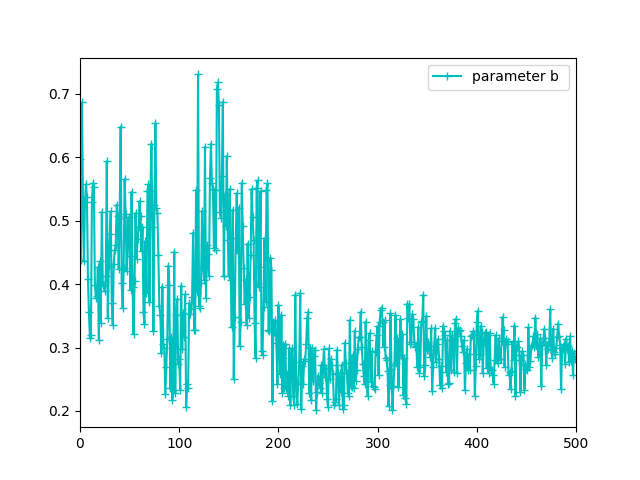} &
\hspace{-0.5cm}\includegraphics[height=4cm,width=4.5cm ]{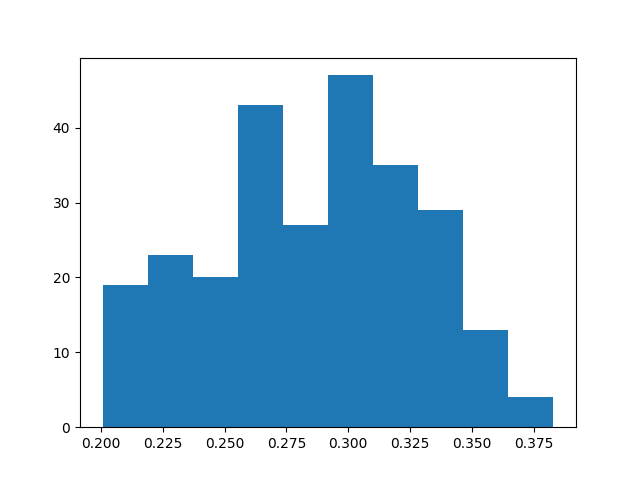} \\
(c) & (d) \\

\hspace{-0.5cm}\includegraphics[height=4cm,width=4.5cm ]{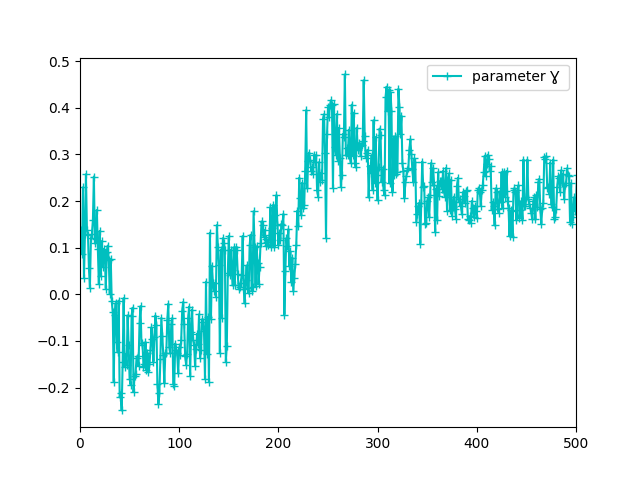} &
\hspace{-0.5cm}\includegraphics[height=4cm,width=4.5cm ]{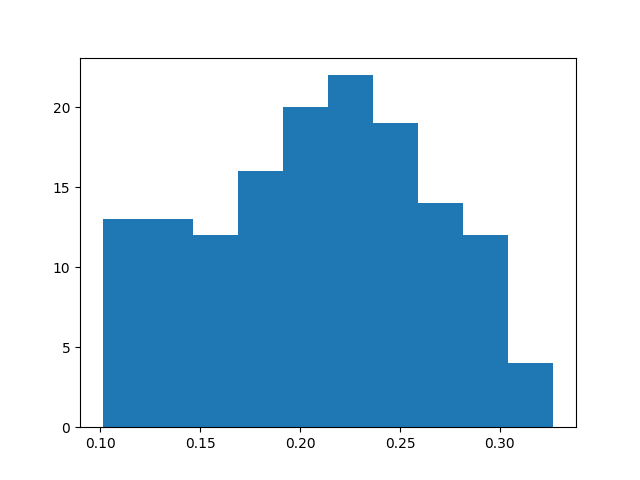} \\
(e) & (f) \\

\end{tabular}
\caption{\centering  Sampling of parameters $c$ (a,b), $b$ (c,d), and $\gamma$ (e,f): chains and histograms. \label{fig:sampled}} 
\end{center}
\end{figure}

After using the proposed Bayesian optimization method to train the CNN models detailed above for the classification of Covid-19 CT images, we analyzed the convergence behavior. Figure \ref{fig:sampled} presents the sampling chains for the $\gamma$, $b$, and $c$ parameters of the proposed MMeLU function (a-c), as well as the histograms of the corresponding samples (d-f). The sampling chains and histograms of the sampled coefficients confirm the good convergence properties of the designed Gibbs sampler. After a burn-in period of 350 iterations, the algorithm achieves stable convergence and exhibits a good mixing rate of the sampled chains.

\subsection{Experiment 1: COVID-19 classification using CT images}
\label{sec:exp_1}
This section examines the effectiveness of our approach in  classifying Covid-19 infections from other pneumonia in CT data.
The COVID-CT dataset\footnote{https://www.kaggle.com/luisblanche/covidct} includes 397 images that are negative for COVID-19 and 349 images that are positive for COVID-19, and belong to 216 patients. We used 566 images for the train and 180 images for the test.\\
Table \ref{tab:exp_1} presents compelling evidence that the proposed Bayesian method outperforms all other activation functions for both the $CNN_1$ and $CNN_2$ architectures. Moreover, except with respect to "ReLU +\cite{fakhfakh2022nonsmooth}", the computational time for convergence is shorter that all the other activation functions. With respect to our previous work ("ReLU +\cite{fakhfakh2022nonsmooth}"), it is worth noting that the proposed method performs better in terms of accuracy and loss value, which confirms the usefulness of the used trainable activation function. However, due to the use of additional parameters, the proposed method is approximately 15$\%$ slower. These conclusions hold for both $CNN_1$ and $CNN_2$.\\
Notably, when regularization is employed, a considerable drop in performance is observed for all competing activation functions, which is largely attributable to the inherent difficulty of classifying CT images due to their content richness and the similarities between Covid-19 infection and other types of pneumonia. \\

\begin{figure*}[!htp]
\begin{center}
\begin{tabular}{ccc}
\includegraphics[height=3.5cm,width=5.5cm ]{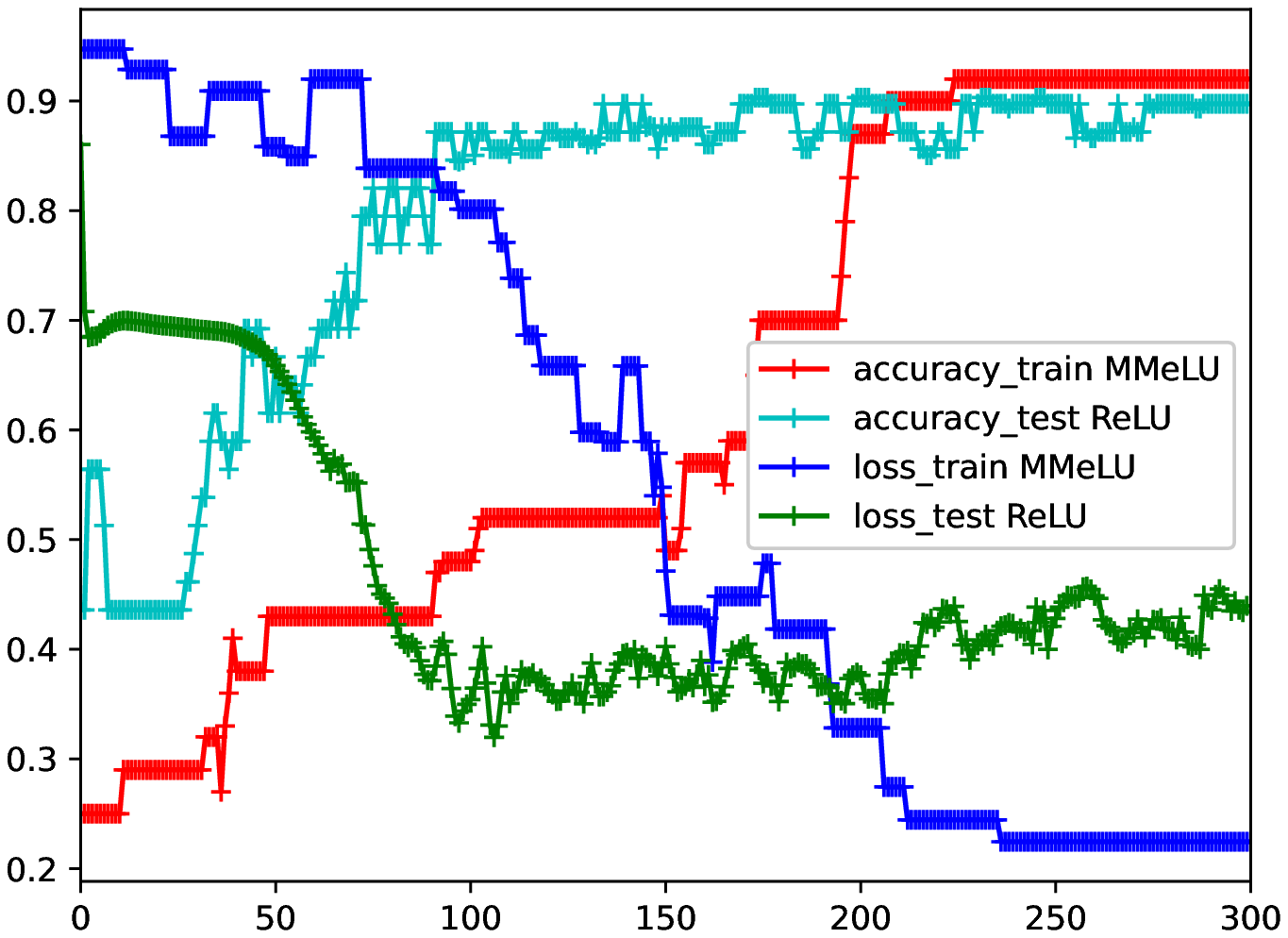}&
\includegraphics[height=3.5cm,width=5.5cm ]{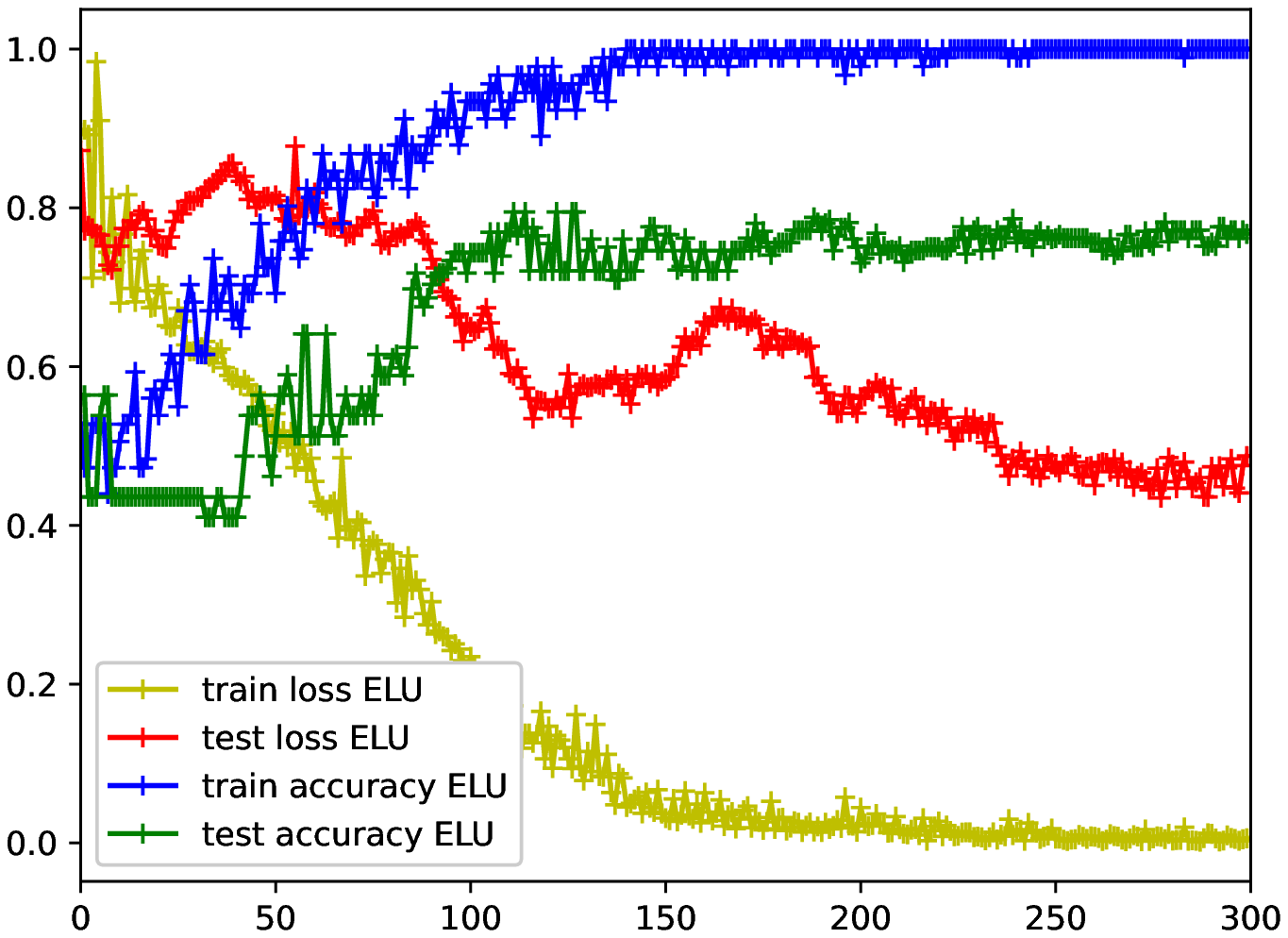}&
\includegraphics[height=3.5cm,width=5.5cm ]{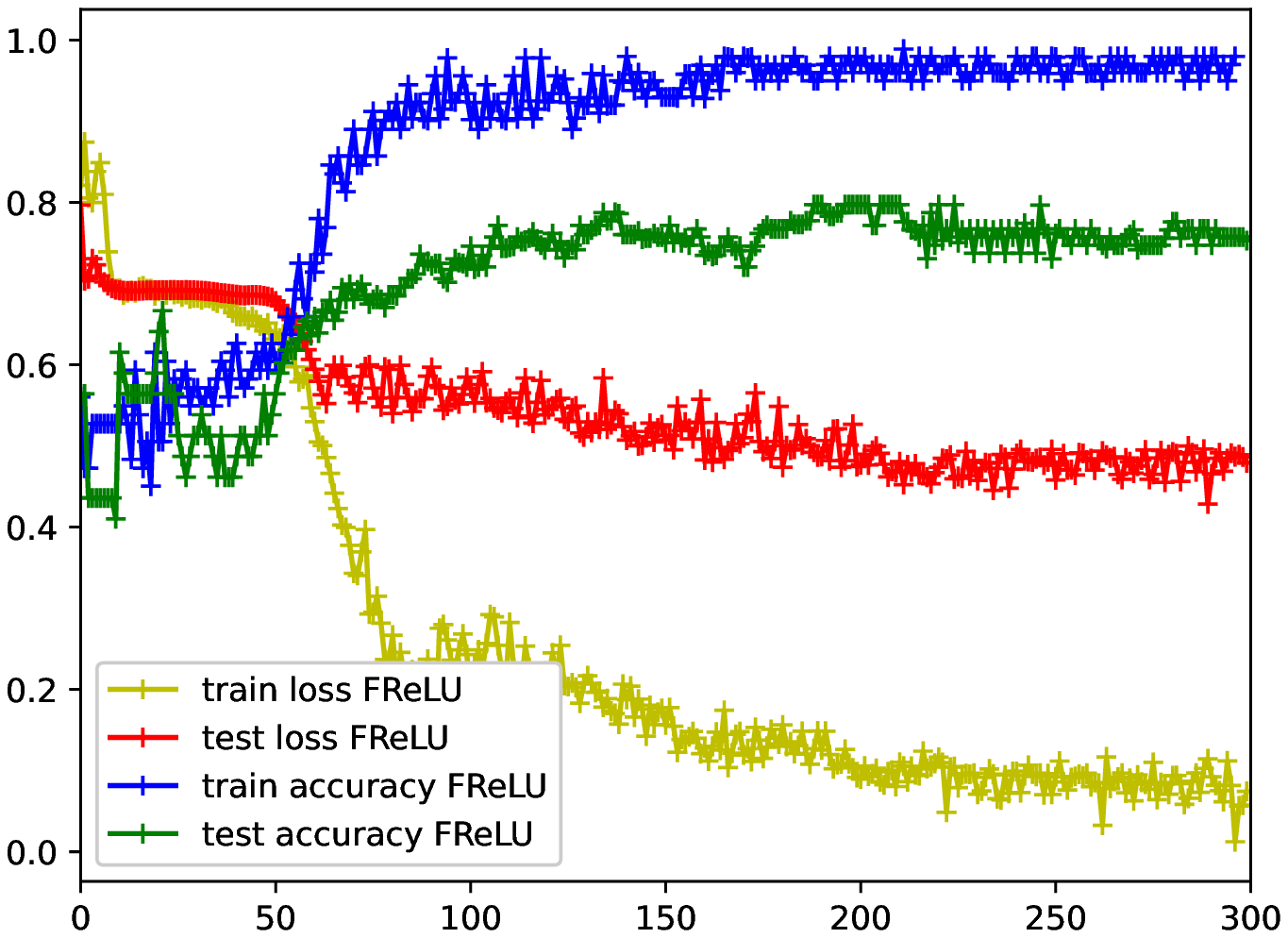}
\\
(a): ReLU & (b): ELU & (c): FReLU \\

\includegraphics[height=3.5cm,width=5.5cm ]{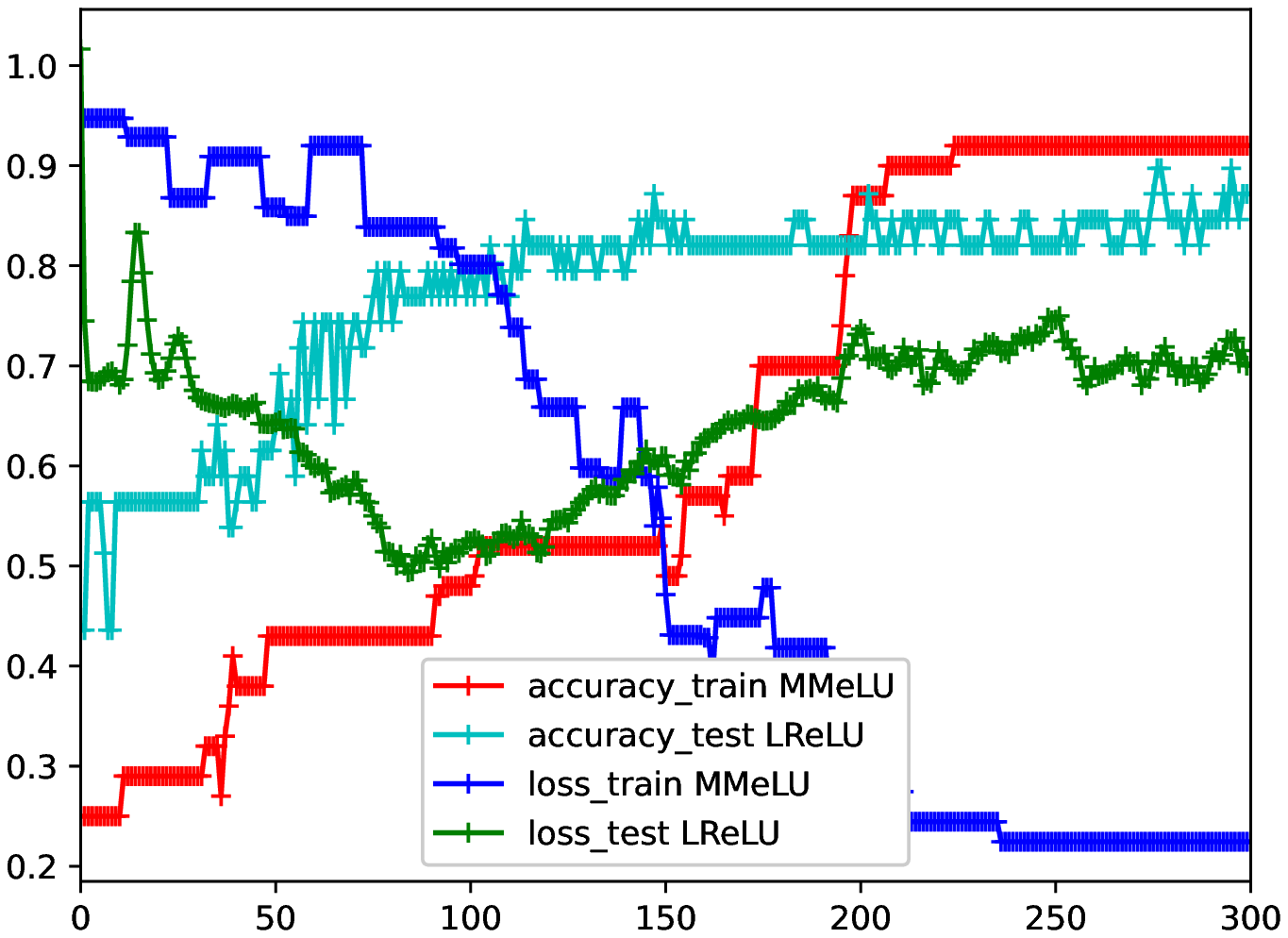}&
\includegraphics[height=3.5cm,width=5.5cm ]{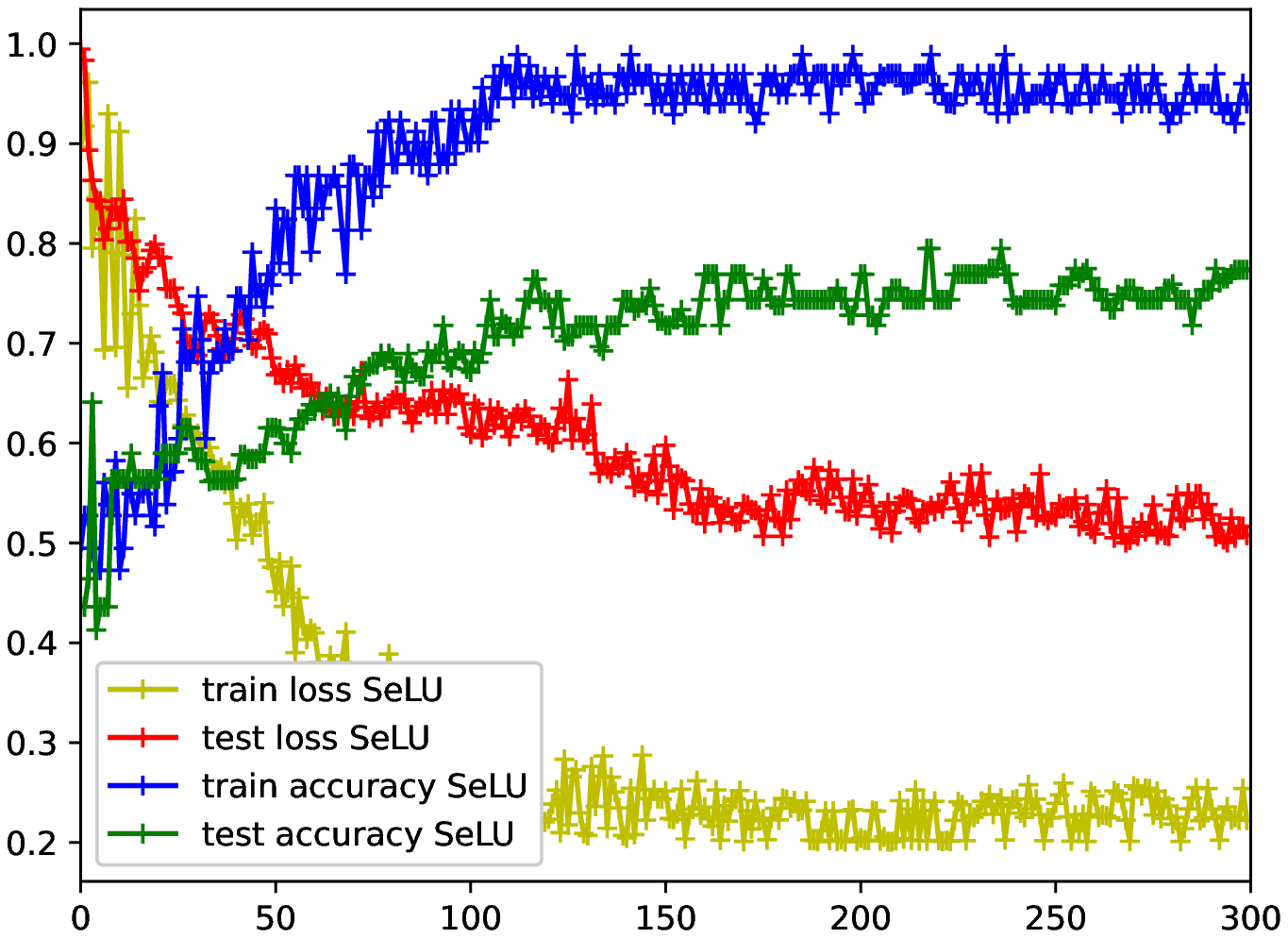}&
\includegraphics[height=3.5cm,width=5.5cm ]{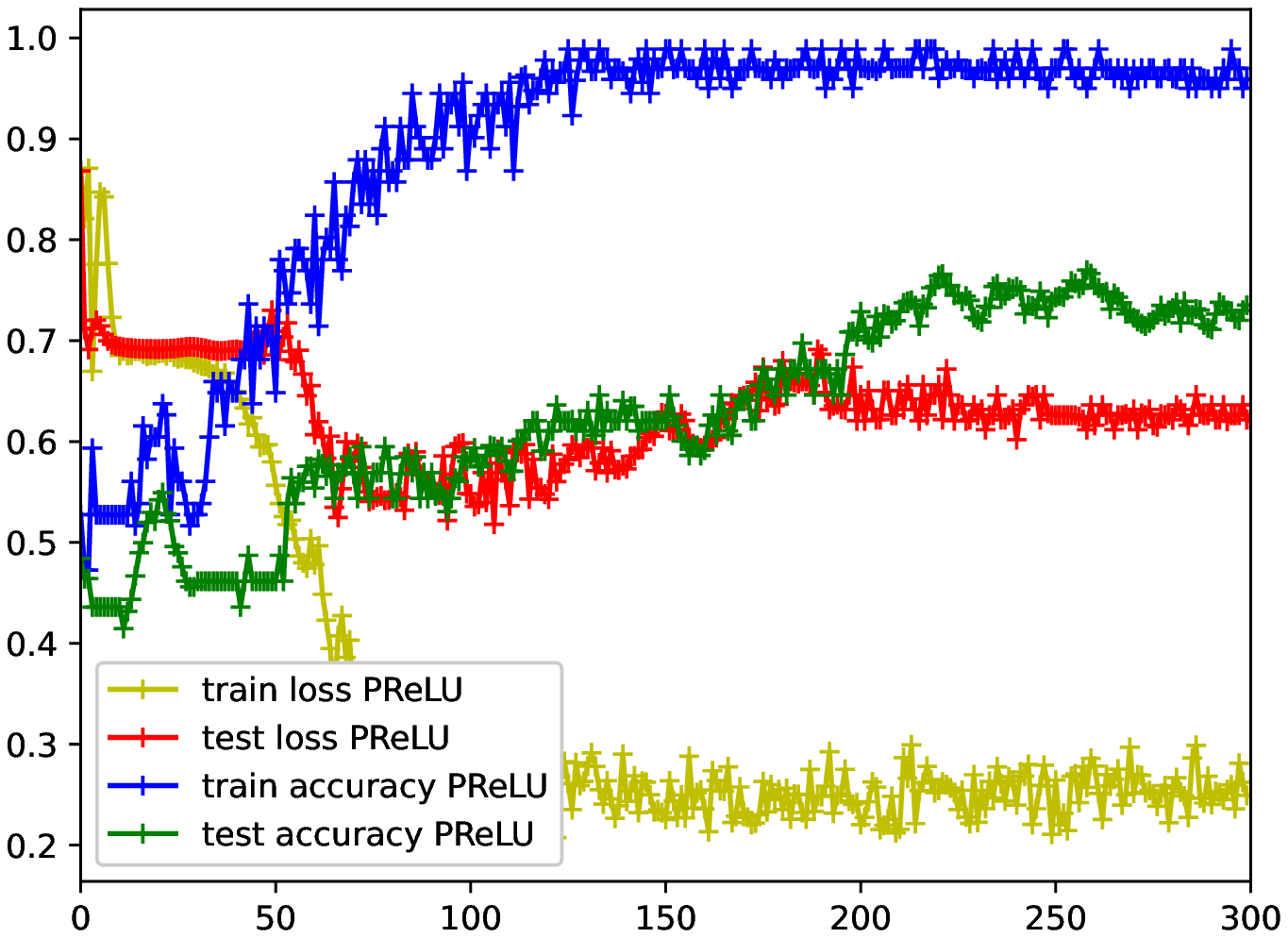}
\\
(d): LReLU & (e): SeLU & (f): PReLU \\

\includegraphics[height=3.5cm,width=5.5cm ]{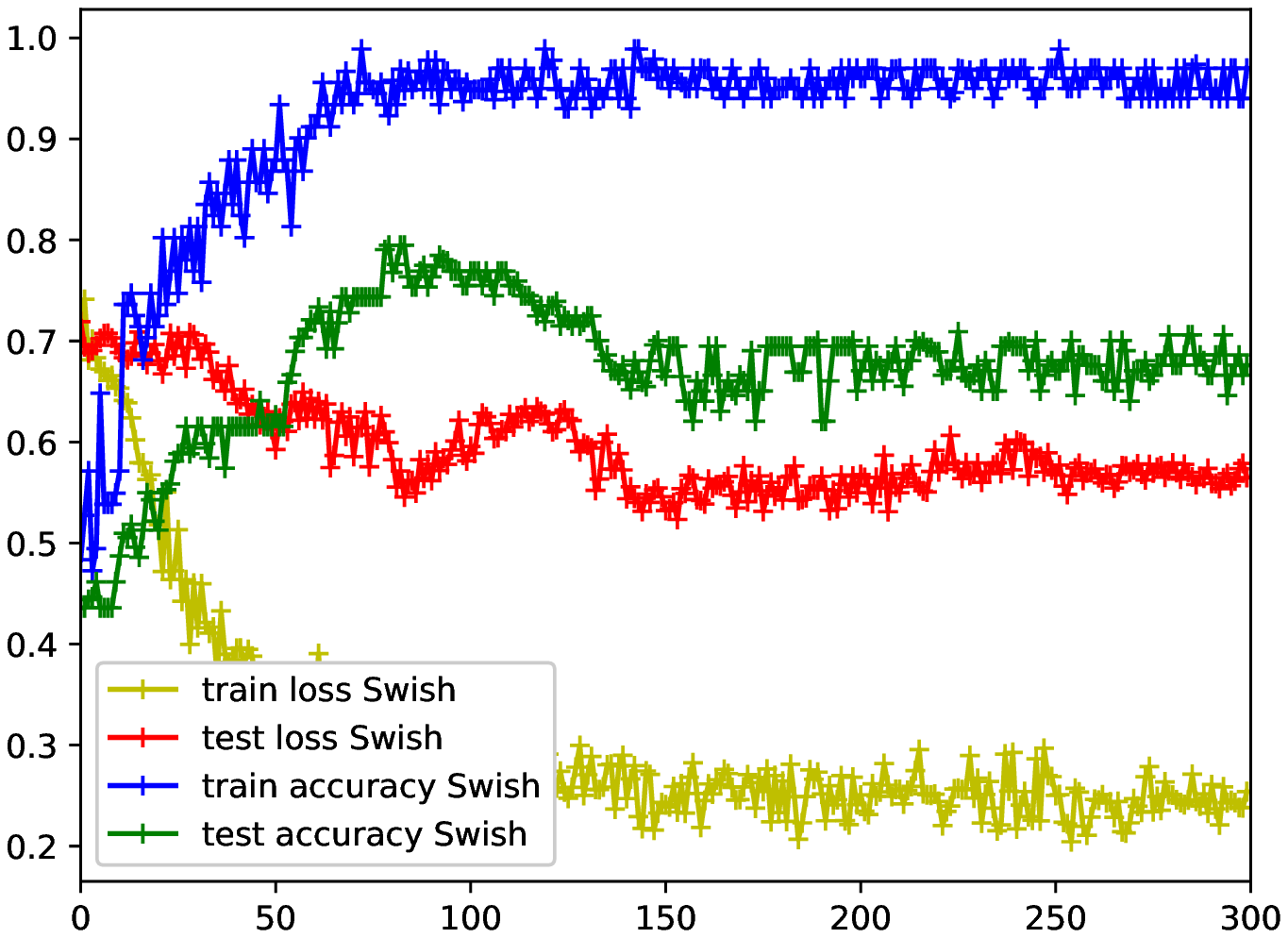}&
\includegraphics[height=3.5cm,width=5.5cm ]{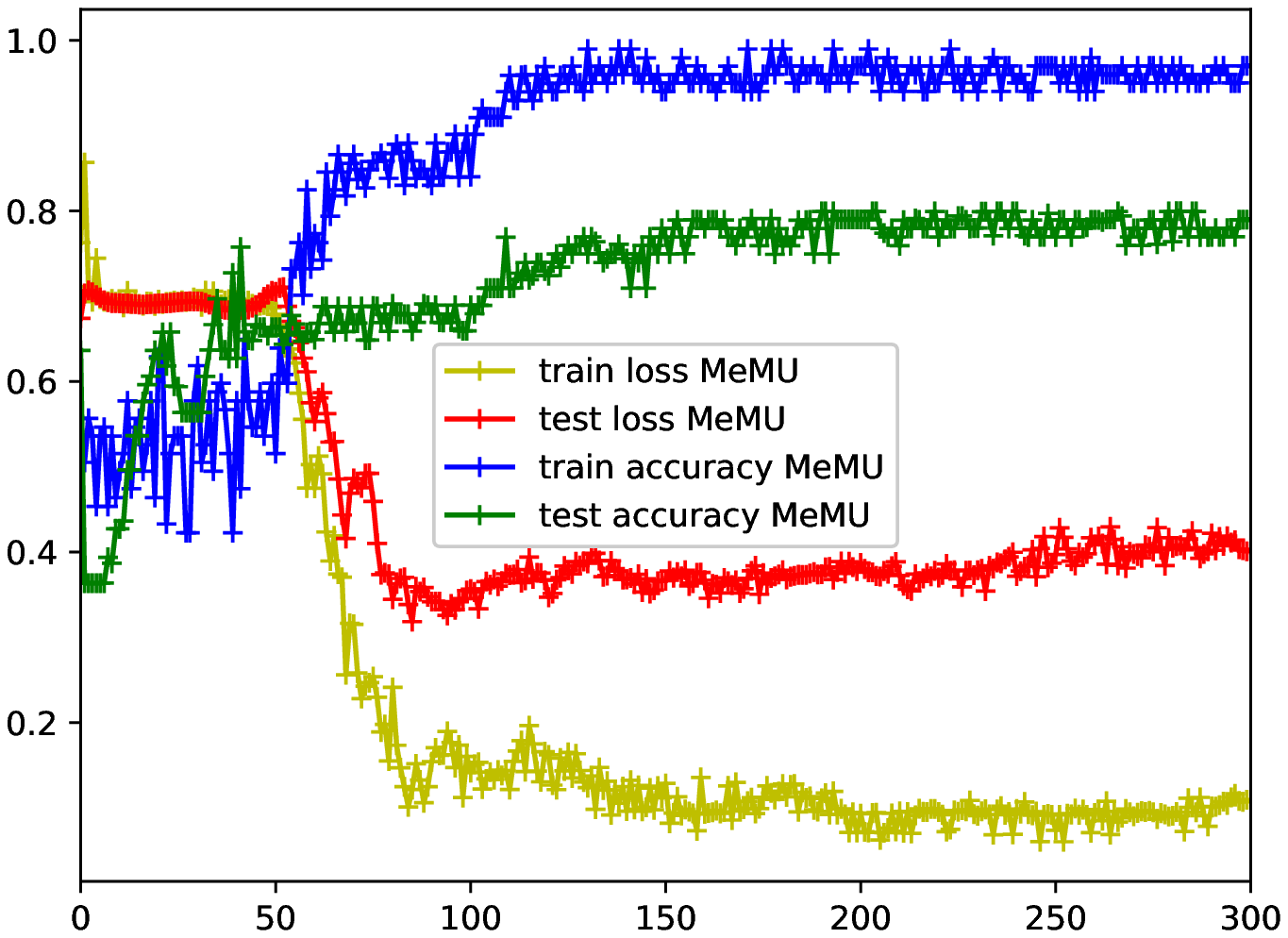}&
\includegraphics[height=3.5cm,width=5.5cm ]{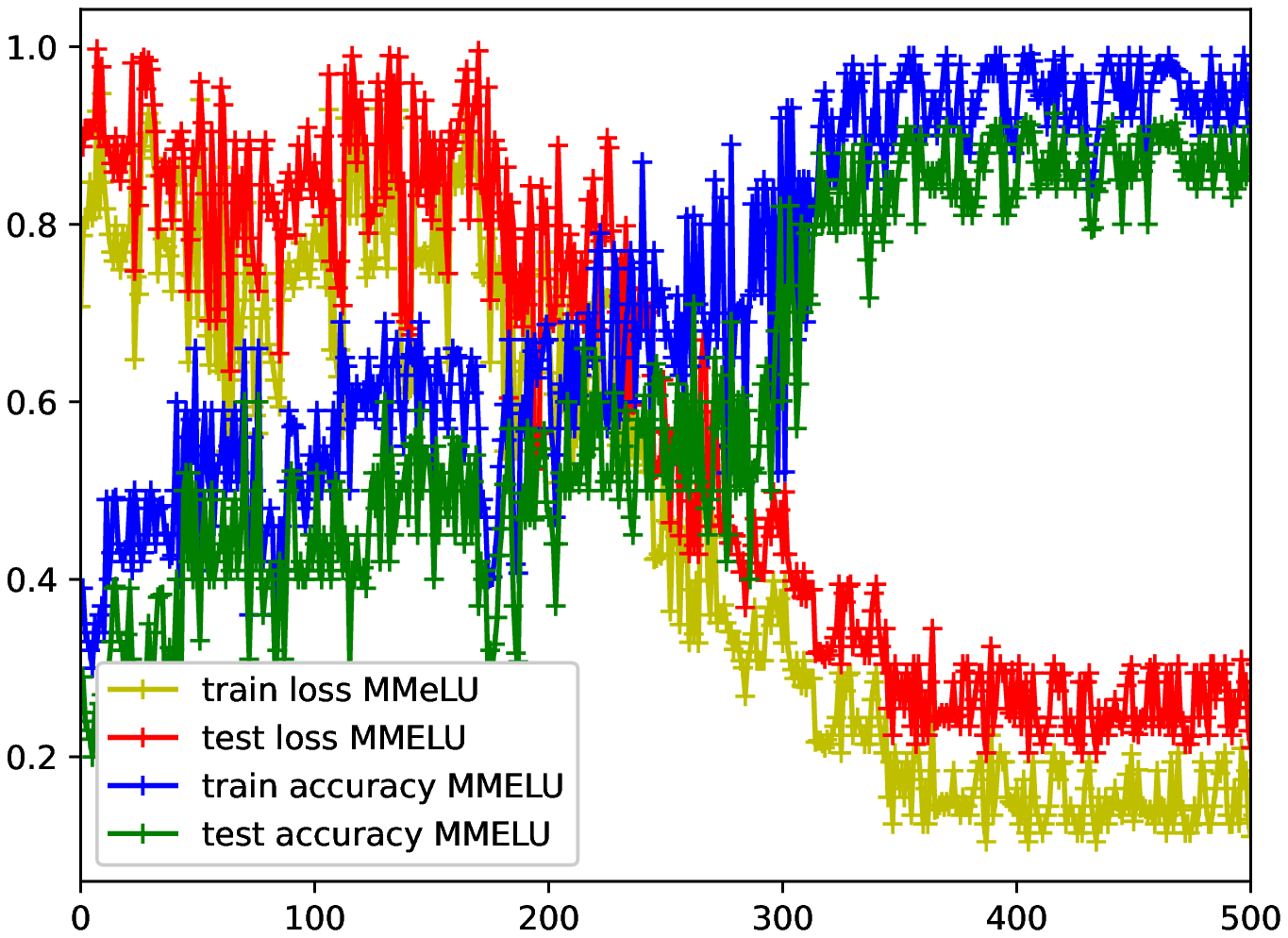}
\\
(g): Swish & (h): MeLU & (i): MMeLU \\

\end{tabular}
\caption{\centering Experiment 1: Train and test curves using $CNN_1$ for all competing activation functions. \label{fig:exp1_CNN_1}} 
\end{center}
\end{figure*}

\begin{figure*}[!htp]
\begin{center}
\begin{tabular}{ccc}
\includegraphics[height=3.5cm,width=5.5cm ]{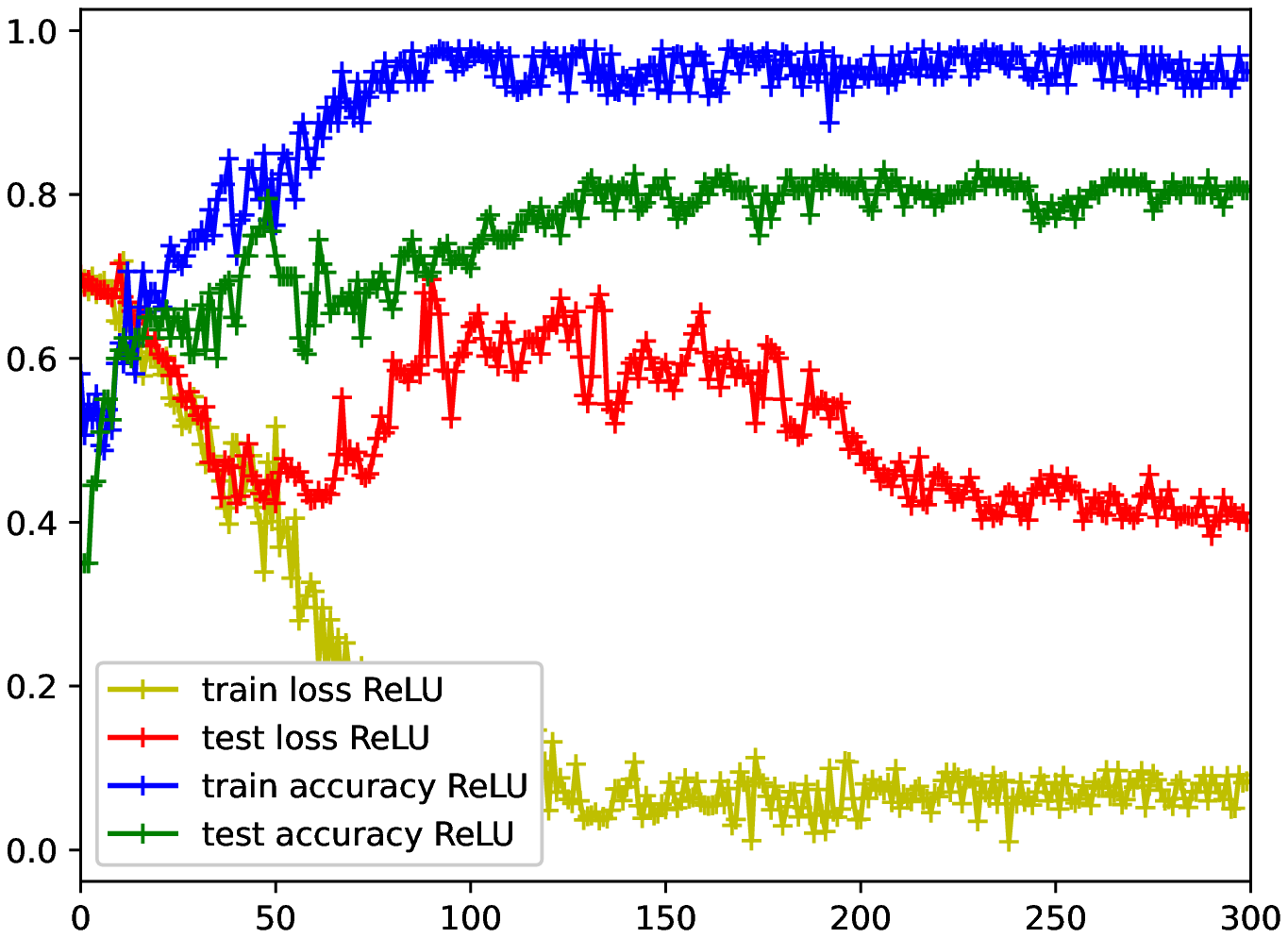}&
\includegraphics[height=3.5cm,width=5.5cm ]{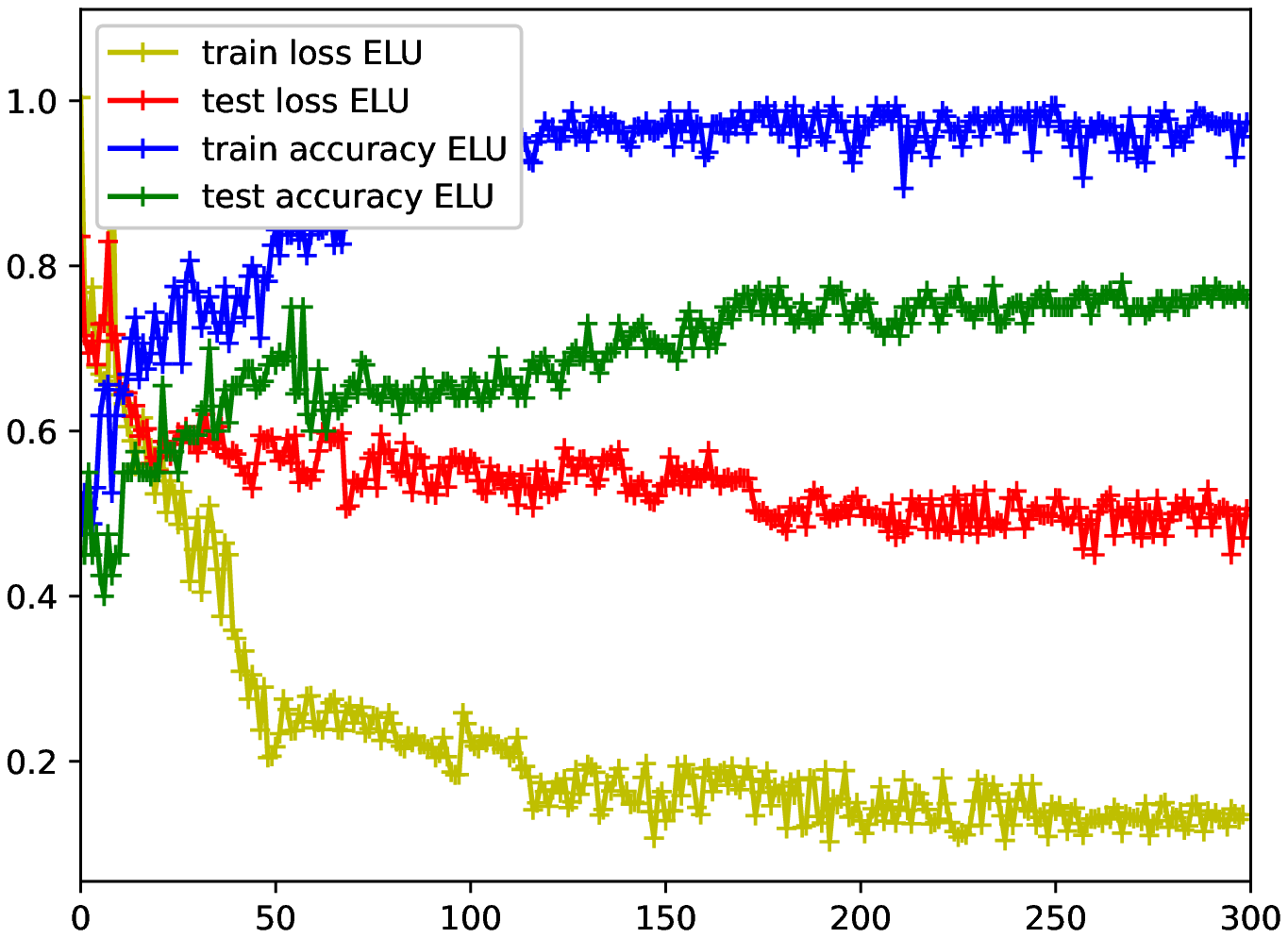}&
\includegraphics[height=3.5cm,width=5.5cm ]{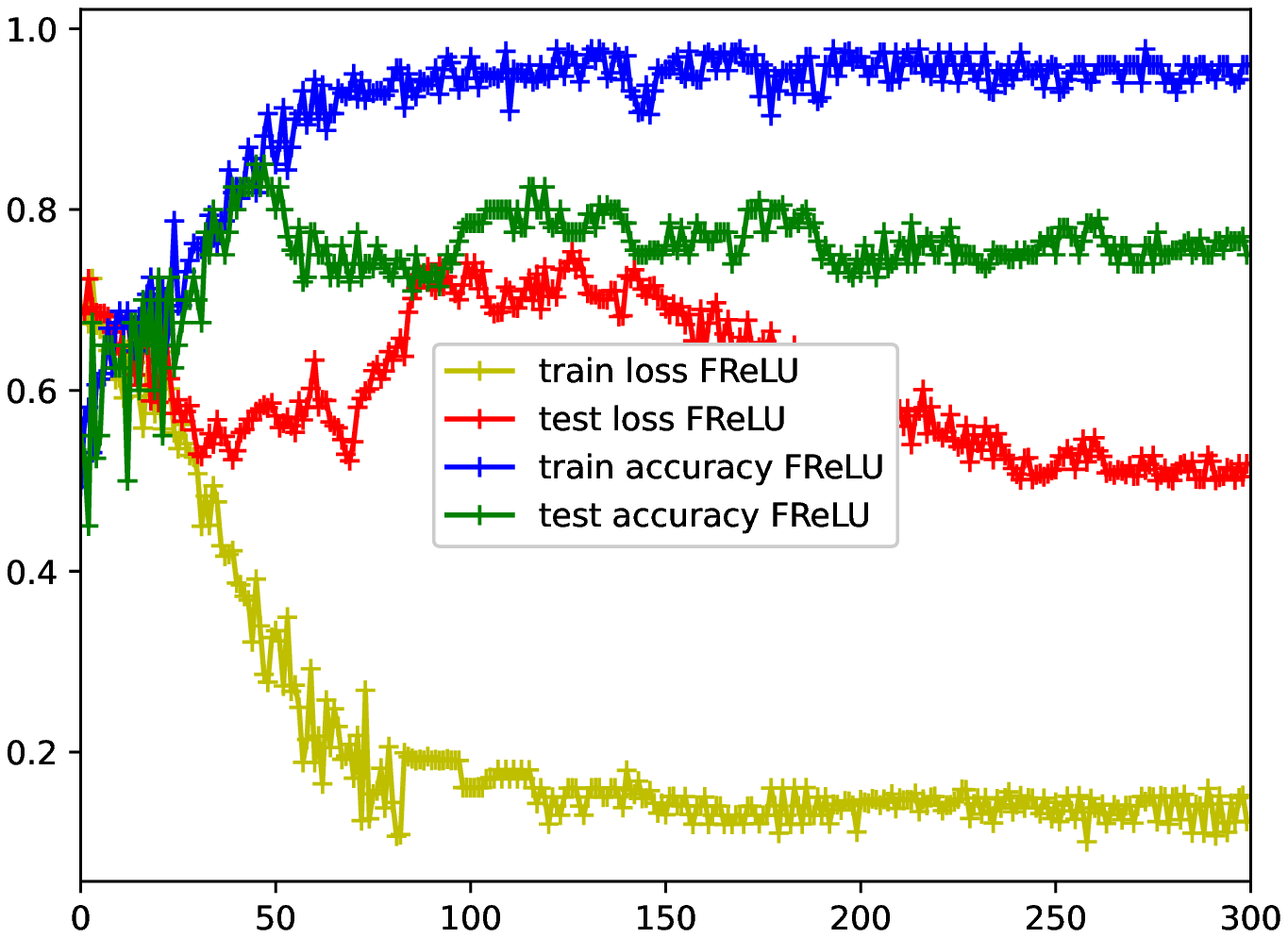}
\\
(a): ReLU & (b): ELU & (c): FReLU \\

\includegraphics[height=3.5cm,width=5.5cm ]{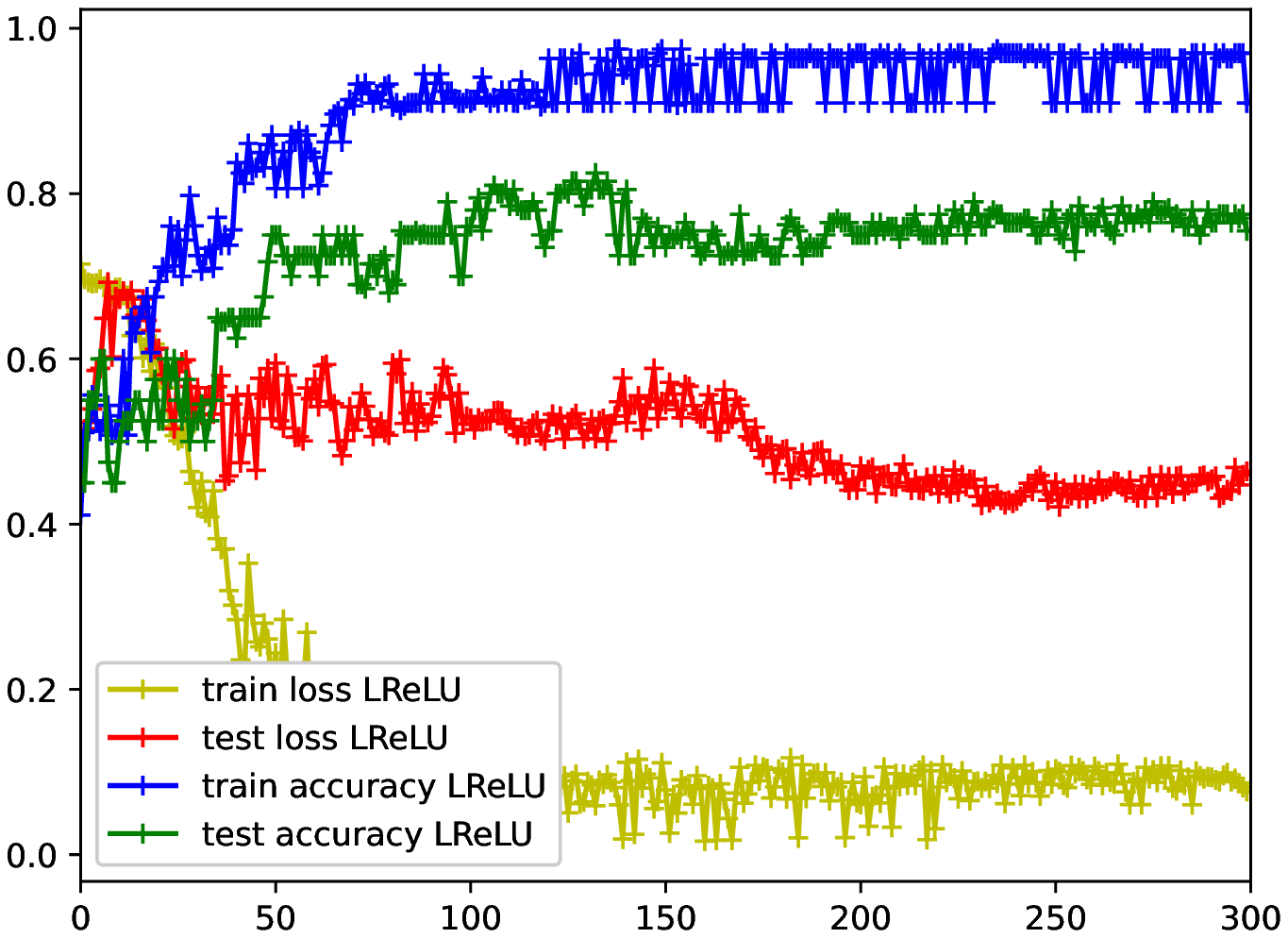}&
\includegraphics[height=3.5cm,width=5.5cm ]{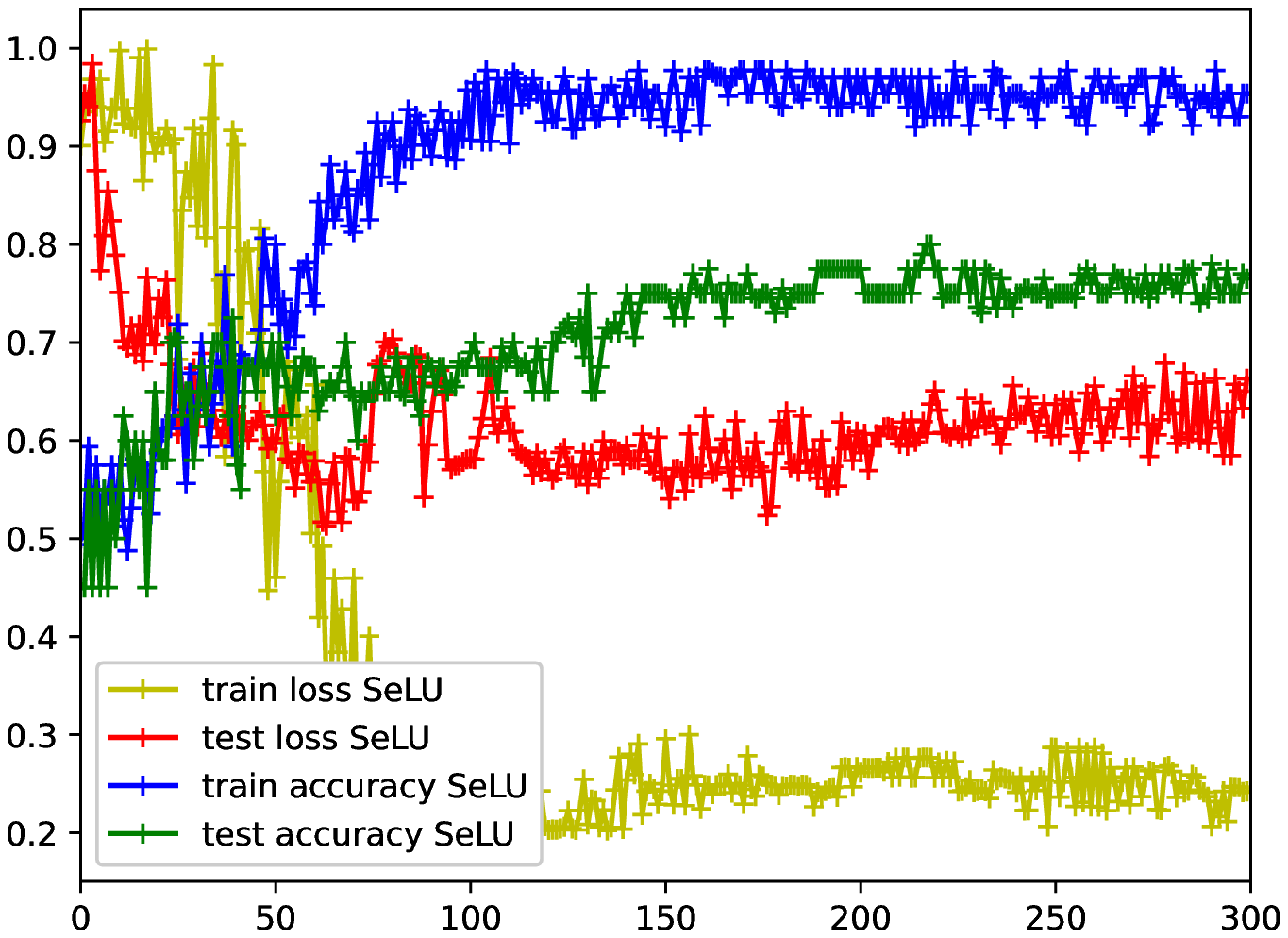}&
\includegraphics[height=3.5cm,width=5.5cm ]{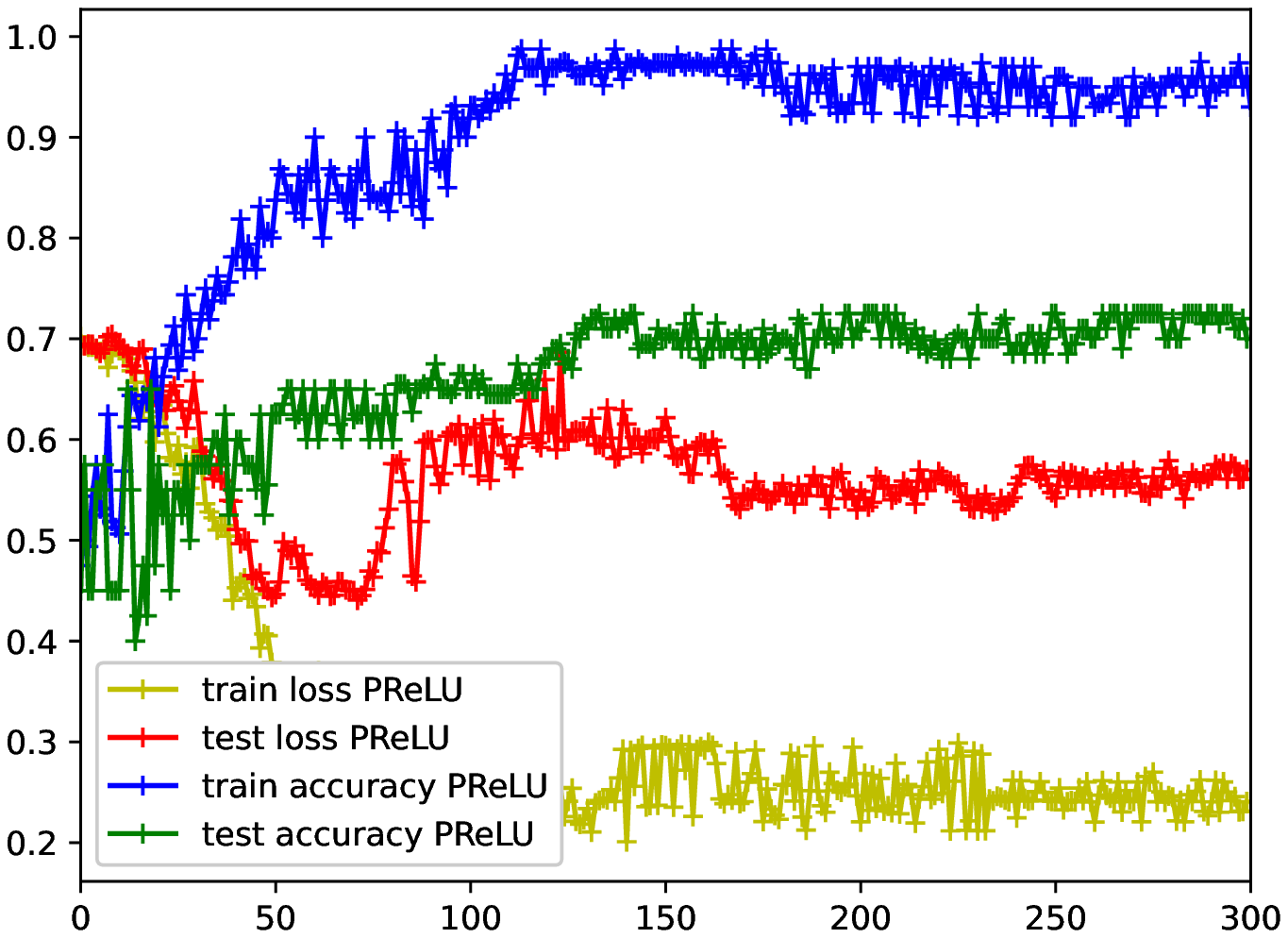}
\\
(d): LReLU & (e): SeLU & (f): PReLU \\

\includegraphics[height=3.5cm,width=5.5cm ]{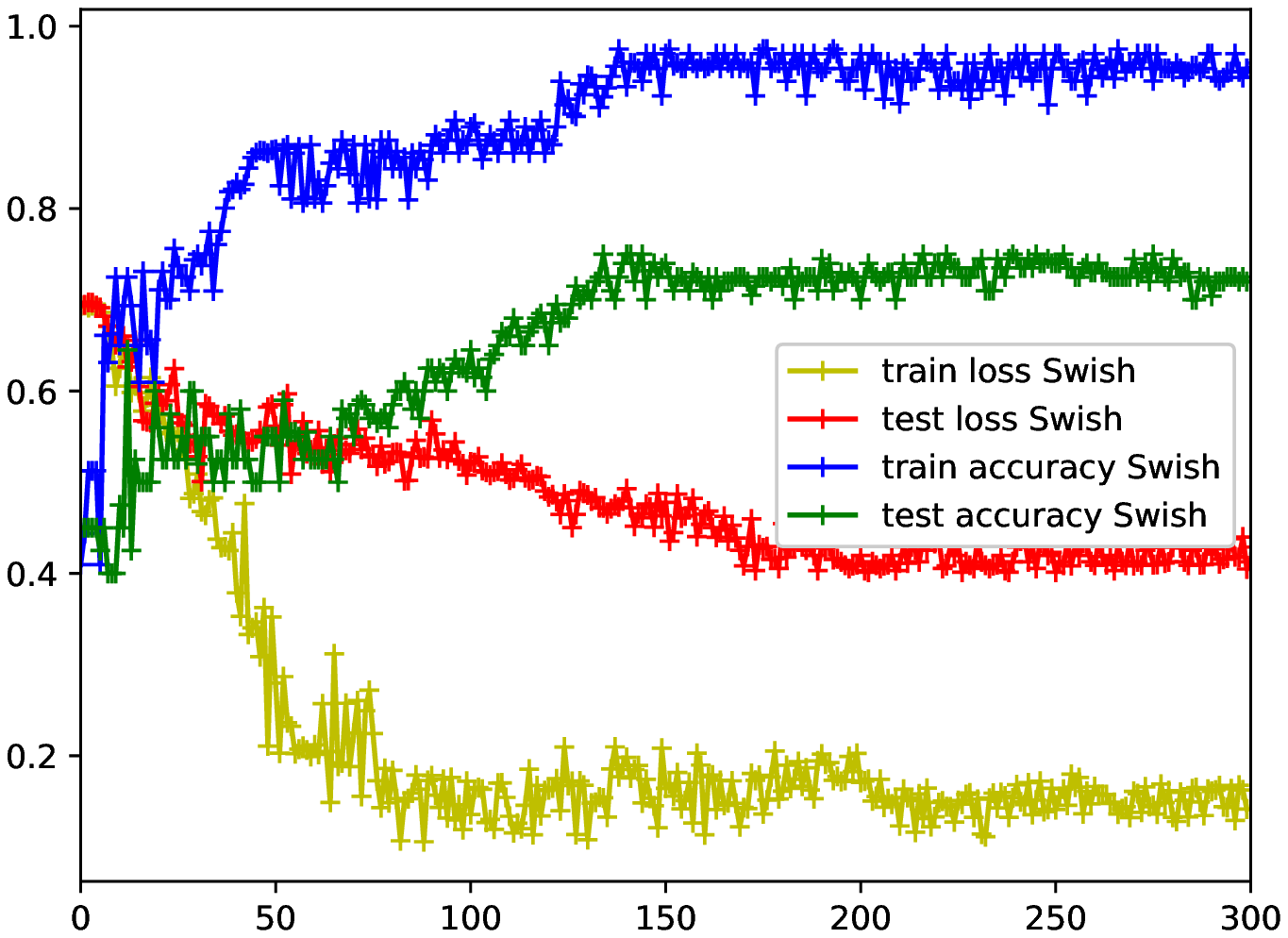}&
\includegraphics[height=3.5cm,width=5.5cm ]{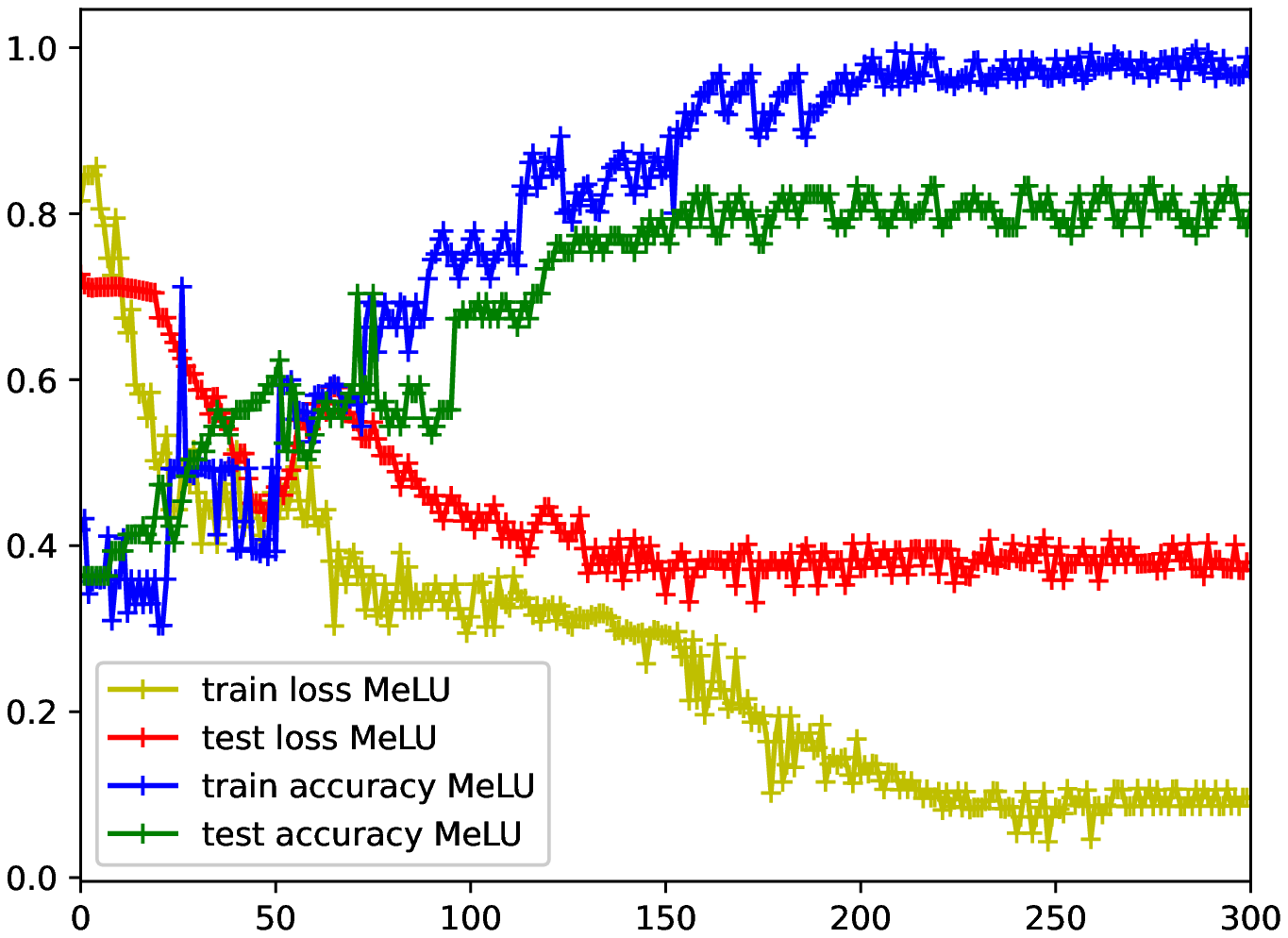}&
\includegraphics[height=3.5cm,width=5.5cm ]{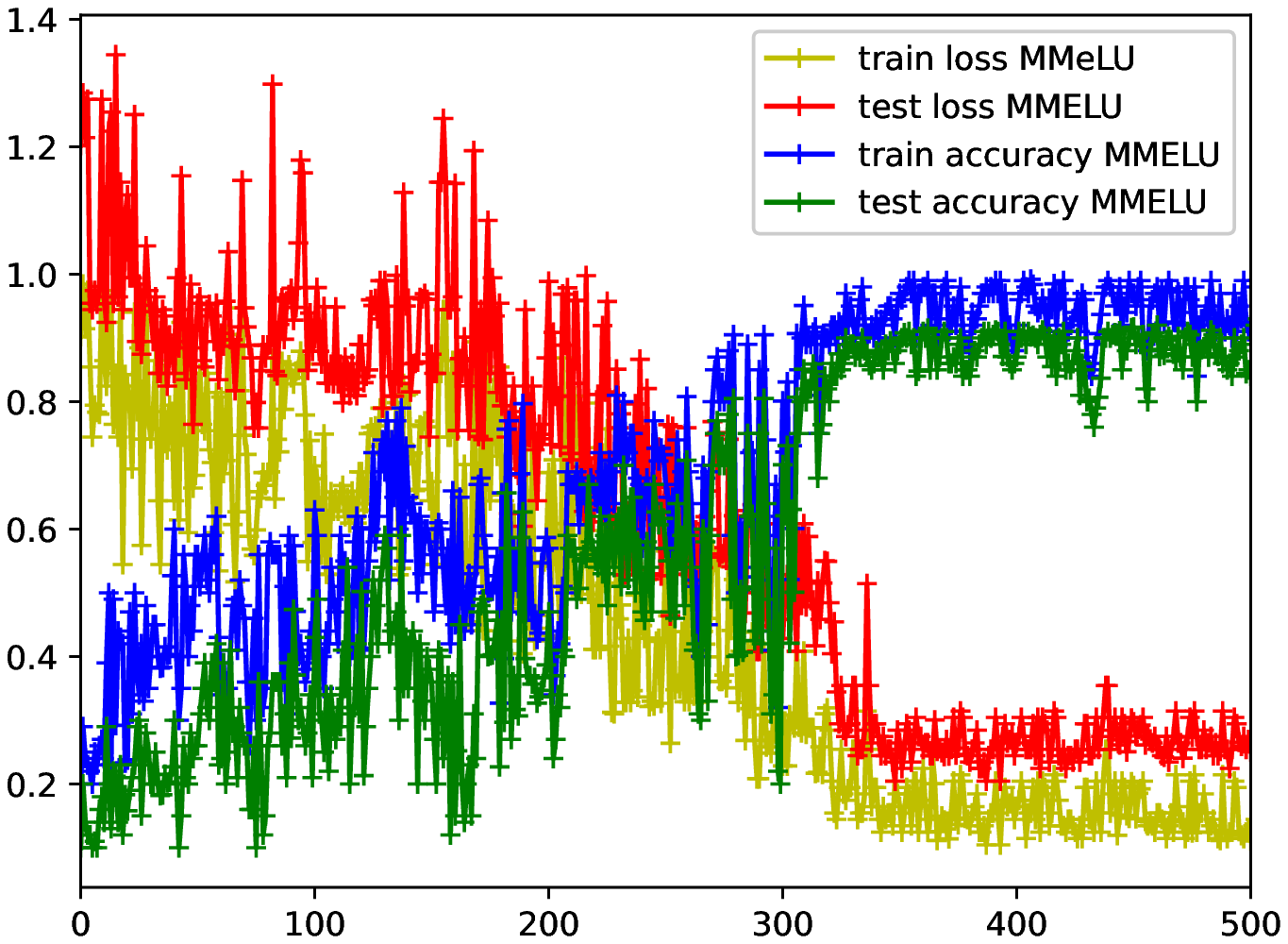}
\\
(g): Swish & (h): MeLU & (i): MMeLU \\

\end{tabular}
\caption{\centering Experiment 1: Train and test curves using $CNN_2$  for all competing activation functions. \label{fig:exp1_CNN_2}} 
\end{center}
\end{figure*}

\begin{table*}[!ht]
\center
\caption
{\footnotesize \label{tab:exp_1} Experiment 1: CT classification results with $CNN_1$ and $CNN_2$ (activation functions (Act Fcts), computation time in minutes, accuracy (Acc), loss, sensitivity (Sens) and specificity (Spec)).}
\begin{tabular}[b]{|l|l|c|c|c|c|c|c|c|c|c|}
\hline
& \multicolumn{5}{|c|}{$CNN_1$} & \multicolumn{5}{|c|}{$CNN_2$}\\
\hline
\hline 
 Act. Fcts & Time & Acc. & Loss &  Sens. & Spec. &  Time & Acc. & Loss &  Sens. & Spec.  \\
\hline 
\hline 
\textbf{MMeLU}  & 46.21 &  \textbf{0.90} & \textbf{0.23} & 
\textbf{0.87} & \textbf{0.86} & 61.92 & \textbf{0.91} & \textbf{0.21} &\textbf{0.87} & \textbf{0.87} \\
\hline 
ReLU+\cite{fakhfakh2022nonsmooth} & \textbf{40} & 0.84 & 0.26 & 0.82 & 0.80 & \textbf{53} & 0.88 & 0.24 & 0.86 & 0.85 \\
\hline 
ReLU  & 58 & 0.73 & 0.43 &  0.69 & 0.68 & 81 & 0.77 & 0.39 & 0.74 & 0.72 \\
\hline 
LReLU  & 65.18 & 0.73 & 0.52 & 0.71 & 0.69 & 105 & 0.78 & 0.44 & 0.76 & 0.75 \\
\hline 
ELU  & 63  & 0.75 & 0.47 &  0.75 & 0.74 & 97  & 0.76 &  0.46 & 0.75 & 0.75 \\
\hline 
PReLU  & 71.28  & 0.68  & 0.72 &  0.64 & 0.62 & 119 & 0.70 &  0.76  & 0.68 & 0.67 \\
\hline 
SeLU  & 64.75  & 0.77  & 0.78 &  0.74 & 0.72 & 107 & 0.76 & 0.69 & 0.74 & 0.73 \\
\hline 
Swish  & 83.41 & 0.68 & 0.58 & 0.65 & 0.62 & 132 & 0.73 & 0.55 & 0.71 & 0.70 \\
\hline 
FReLU  & 77.8  & 0.76  & 0.59 &  0.76 & 0.75 & 123 & 0.77 & 0.52  &  0.76 & 0.75 \\
\hline 
MeLU  & 95.89  & 0.77 & 0.43 &  0.77 & 0.76 & 146 & 0.80 & 0.38 & 0.80 & 0.80 \\
\hline
\end{tabular}
\end{table*}
Learning and test curves (accuracy and loss) illustrated in Figures \ref{fig:exp1_CNN_1} and \ref{fig:exp1_CNN_2} confirm the good behavior of the proposed method, which is not necessarily the case of the other competing models where a marked difference between the precision and loss curves can be noticed. For example, while the LReLU function introduces a negative bias that suppresses excessive activations, an inappropriate bias value can lead to underfitting. Similarly, although the ELU function allows  negative activation, its exponential form can lead to an explosion of the activation value for large values of $x$.\\
The Swish function is known to accelerate learning convergence, but it can also lead to overfitting by being more sensitive to outliers. Likewise, although the FReLU function can capture complex data patterns, it can also suffer from overfitting if the parameters are not well chosen. \\
These remarkable differences confirm the interest  and efficiency of our MMeLU function, which outperforms all competing activation functions in terms of accuracy and robustness to regularization.

\subsection{Experiment 2: Fashion-MNIST image classification}
\label{sec:exp_2}
The learning performance of the competing activation function algorithms is assessed in this scenario using the standard \textit{Fashion-MNIST} dataset. A training set of 60,000 images is employed, with a test set of 10,000 images. Each example is a $28 \time 28$ grayscale image paired with a label from one of ten classes, with 7,000 images in each class. We used 48,000 images for the train set and 12,000 for the test set for model training. \\
Table \ref{tab:exp_2} reports the results obtained for the \textit{Fashion-MNIST} dataset. Our proposed method outperformed all other competing activation functions, achieving a minimum accuracy of 93\% for both CNN models. 
Additionally, Table \ref{tab:exp_2} shows that the processing time for competing methods is greater than 150 minutes for $CNN_1$, while only almost 112 minutes are required for our method. Similar conclusions hold with $CNN_2$. \\
Moreover, our proposed MMeLU function showed similar global performance compared to "ReLU + \cite{fakhfakh2022nonsmooth}" for the two CNN models. One more advantage of MMeLU is its fully automatic estimation of parameters, which is not the case for ReLU + \cite{fakhfakh2022nonsmooth}, where some parameters need to be manually set. Furthermore, our proposed method has demonstrated its ability to achieve remarkable performances in challenging cases, as shown in the first experiment.

\begin{table*}[!ht]
\center
\caption
{\footnotesize \label{tab:exp_2} Experiment 2: Fashion-MNIST classification results with $CNN_1$ and $CNN_2$ (activation functions (Act Fcts), computation time in minutes, accuracy (Acc), loss, sensitivity (Sens) and specificity (Spec)).}
\begin{tabular}[b]{|l|l|c|c|c|c|c|c|c|c|c|}
\hline
& \multicolumn{5}{|c|}{$CNN_1$} & \multicolumn{5}{|c|}{$CNN_2$}\\
\hline
\hline 
 Act. Fcts& Time & Acc. & Loss &  Sens. & Spec. &  Time & Acc. & Loss &  Sens. & Spec.  \\
\hline 
\hline 
\textbf{MMeLU}  & 111.6 & \textbf{0.93} & \textbf{0.20} & \textbf{0.90} & \textbf{0.89} & 319.8 & \textbf{0.94} & \textbf{0.19} & \textbf{0.92} &\textbf{0.91} \\
\hline 
ReLU+\cite{fakhfakh2022nonsmooth} & \textbf{90.5} & 0.92 & 0.22 & \textbf{0.90} & 0.88 &\textbf{308}.4 & 0.93 & \textbf{0.19} & 0.91 & 0.89 \\
\hline 
ReLU  & 145.4 & 0.90 & 0.38 & 0.85 & 0.82 & 421.3 & 0.91 & 0.32 & 0.88 & 0.87  \\
\hline 
LReLU  & 157.2 & 0.86 & 0.36 & 0.85 & 0.83 & 409.7 & 0.87 & 0.34 & 0.85 & 0.84 \\
\hline 
ELU  & 154.5 & 0.87 & 0.33 & 0.86 & 0.84 & 389 & 0.88 & 0.32 & 0.86 & 0.85 \\
\hline 
PReLU  & 158.7 & 0.87 & 0.32 & 0.86 & 0.85 & 400.3 & 0.85 & 0.33 & 0.83 & 0.82 \\
\hline 
SeLU  & 150 & 0.85 & 0.34 & 0.84 & 0.83 & 383.8 & 0.87 & 0.33 & 0.86 & 0.85 \\
\hline 
Swish  & 148.5 & 0.86 & 0.36 & 0.83 & 0.81 & 423 & 0.87 & 0.35 & 0.84 & 0.82 \\
\hline 
FReLU  & 149.8 & 0.87 & 0.35 & 0.85 & 0.83 & 395.7 & 0.86 & 0.34 & 0.85 & 0.84 \\
\hline 
MeLU  & 169 & 0.89 & 0.33 & 0.86 & 0.84 & 452.5 & 0.89 & 0.31 & 0.88 & 0.87 \\
\hline
\end{tabular}
\end{table*}

\subsection{Experiment 3: CIFAR-10 image classification}
\label{sec:exp_3}
In this experiment, the learning performance of the proposed model (made up of the trainable activation function and the Bayesian sparse optimization) is evaluated using the standard \textit{CIFAR-10} dataset. The CIFAR-10 dataset contains 60000 $32 \times 32$ color images divided into 10 classes with 6000 images per class. There are 50,000 training and 10,000 test images. \\
Table \ref{tab:exp_3} presents the classification results for the \textit{CIFAR-10} dataset. It can be observed that the proposed Bayesian model performed well overall, even when multiple classes were used. In contrast, all competing activation functions (with Adam optimizer) had similar accuracy, around 88\%, with a loss rate almost double that of the proposed model for both architectures. \\ 
Moreover, the MMeLU activation function takes significantly less time for training than other activation functions + Adam, which often take a long time before convergence. For the competing functions when used with Adam, we noticed similar performance between the trainable and standard activation functions, with a slight superiority observed with the trainable function ELU.\\
The same conclusions can be drawn by examining the results of "ReLU + \cite{fakhfakh2022bayesian}" on this dataset. These findings suggest that the proposed MMeLU function provides better flexibility for the activation task, even for the multi-class case.


\begin{table*}[!ht]
\center
\caption
{\footnotesize \label{tab:exp_3} Experiment 3: CIFAR-10 classification results with $CNN_1$ and $CNN_2$ (activation functions (Act Fcts), computation time in minutes, accuracy (Acc), loss, sensitivity (Sens) and specificity (Spec)).}
\begin{tabular}[b]{|l|l|c|c|c|c|c|c|c|c|c|}
\hline
& \multicolumn{5}{|c|}{$CNN_1$} & \multicolumn{5}{|c|}{$CNN_2$}\\
\hline 
Act. Fcts & Time & Acc. & Loss &  Sens. & Spec. &  Time & Acc. & Loss &  Sens. & Spec.  \\
\hline 
\textbf{MMeLU} & 120.7 & \textbf{0.91} & \textbf{0.21}  & \textbf{0.89} & \textbf{0.88} & 332.5 & \textbf{0.93} & \textbf{0.20} &\textbf{0.90} & \textbf{0.88} \\
\hline 
ReLU+\cite{fakhfakh2022nonsmooth} & \textbf{100.7} & 0.90 & 0.25 & \textbf{0.89} & 0.87 & \textbf{331} & 0.92 & 0.21 & \textbf{0.90} & 0.87 \\
\hline 
ReLU  & 161 & 0.87 & 0.42 & 0.83 & 0.81 & 429 & 0.90 & 0.36 & 0.87 & 0.86 \\
\hline 
LReLU  & 170 & 0.87 & 0.52 & 0.85 & 0.84 & 437.9 & 0.85 & 0.55 & 0.83 & 0.81 \\
\hline 
ELU  & 165 & 0.88 & 0.41 & 0.86 & 0.85 & 438 & 0.88 & 0.39 & 0.86 & 0.85 \\
\hline 
PReLU  & 198.2 & 0.82 & 0.48 & 0.84 & 0.82 & 466.8 & 0.84 & 0.47 & 0.81 & 0.79 \\
\hline 
SeLU  & 173.6 &  0.84 & 0.43 & 0.85 & 0.84 & 454.3 & 0.86 & 0.39 & 0.87 & 0.86 \\
\hline 
Swish  & 209.7 & 0.83 & 0.49 & 0.82 & 0.79 & 478.3 & 0.85 & 0.46 & 0.83 & 0.81 \\
\hline 
FReLU  & 172.3 &  0.84 & 0.41 & 0.82 & 0.81 & 431.7 & 0.86 & 0.38 & 0.83 & 0.81 \\
\hline 
MeLU  & 219.9 & 0.86 & 0.40 & 0.83 & 0.83 & 485.6 & 0.90 & 0.34 & 0.88 & 0.87 \\
\hline
\end{tabular}
\end{table*}

\subsection{Experiment 4: Comparison on Deep CNN}
\label{sec:deep_cnns}
This section investigates the performance of our algorithm on the standard Fashion-MNIST dataset using a deep CNN.
This CNN is claimed to be deeper than $CNN 1$ and $CNN 2$.
It has 25 convolutional layers (5 X Conv3x3-32, 5 X Conv3x3-64, 5 X Conv3x3-128, 5 X Conv3x3-256 and 5 X Conv3x3-512) and four layers (FC-512, FC-256, Fc-128 and FC-softmax).
All of them use convolutional layers with $3 \times 3$ kernel filters as well as $2 \times 2$ max-pooling and stride size of 1. \\
The results demonstrate that our proposed MMeLU method maintains good performance on this deep architecture, as shown in table \ref{tab:6_exp_4}. The same conclusions can be drawn as from previous experiments regarding the superiority of MMeLU over competing activation functions, including our previous work in \cite{fakhfakh2022nonsmooth}. Interestingly, in addition to better scores in comparison to \cite{fakhfakh2022nonsmooth}, the proposed method provides convergence time which is only 2 $\%$ slower. It turns out that when the depth of the network increases (similarly the number of parameters), the additional cost to estimate the activation function parameters is no longer significant while reaching better performance. \\

\begin{table}[!ht]
\center
\caption
{\footnotesize \label{tab:6_exp_4} Results of Fashion-MNIST classification with a deep CNN (activation functions (Act Fcts), computation time in minutes, accuracy, loss, sensitivity (Sens), and specificity (Spec) ).}
\begin{tabular}[b]{|l|l|c|c|c|c|c|c|}
\hline
 Act. Fcts& Time (min) & Accuracy & Loss & Sens  & Spec  \\
\hline
\hline
\textbf{MMeLU} & 651.4 & \textbf{0.94} & \textbf{0.19} & \textbf{0.92} & \textbf{0.92} \\
\hline 
ReLU+\cite{fakhfakh2022nonsmooth} & \textbf{638} & 0.93 & 0.21 & 0.91 & 0.90 \\
\hline 
ReLU  & 854 & 0.89 & 0.34 & 0.87 & 0.85 \\
\hline 
LReLU  & 847.9 & 0.86 & 0.38 & 0.84 & 0.82 \\

\hline 
ELU  & 870 & 0.87 & 0.35 & 0.85 & 0.84 \\
\hline 
PReLU  & 855.6 & 0.86 & 0.36 & 0.83 & 0.83 \\
\hline 
SeLU  & 867.3 & 0.88 & 0.33 & 0.87 & 0.86 \\
\hline 
Swish  & 880 & 0.87 & 0.34 & 0.86 & 0.85 \\
\hline 
FReLU  & 873.4 & 0.85 & 0.36 & 0.84 & 0.82 \\
\hline 
MeLU  & 890 & 0.90 & 0.32 & 0.86 & 0.85 \\
\hline
\end{tabular}
\end{table}

\section{Conclusion}
\label{sec:conlusion}
In this paper, we proposed a Bayesian approach for training sparse deep neural networks with trainable activation functions. Our method learns the weights and parameters of the proposed activation function directly from the data without any user configuration. Compared to competing algorithms, our approach achieves promising results with better classification accuracy and high generalization performance, while also being computationally efficient, particularly in challenging cases. Our experiments on standard datasets such as Fashion-MNIST and CIFAR-10 demonstrate the precision and stability of our approach, as well as its general applicability. \\
Future work will focus on parallelizing the algorithm to enable GPU calculations and further reduce computational time.

\bibliographystyle{elsarticle-num}
\bibliography{references}

\begin{thebibliography}{10}
\expandafter\ifx\csname url\endcsname\relax
  \def\url#1{\texttt{#1}}\fi
\expandafter\ifx\csname urlprefix\endcsname\relax\def\urlprefix{URL }\fi
\expandafter\ifx\csname href\endcsname\relax
  \def\href#1#2{#2} \def\path#1{#1}\fi

\bibitem{alsarhan2021machine}
A.~Alsarhan, M.~Alauthman, E.~Alshdaifat, A.-R. Al-Ghuwairi, A.~Al-Dubai,
  Machine learning-driven optimization for svm-based intrusion detection system
  in vehicular ad hoc networks, Journal of Ambient Intelligence and Humanized
  Computing (2021) 1--10.

\bibitem{wang2020deep}
Z.~Wang, J.~Chen, S.~C. Hoi, Deep learning for image super-resolution: A
  survey, IEEE transactions on pattern analysis and machine intelligence
  43~(10) (2020) 3365--3387.

\bibitem{dong2015image}
C.~Dong, C.~C. Loy, K.~He, X.~Tang, Image super-resolution using deep
  convolutional networks, IEEE transactions on pattern analysis and machine
  intelligence 38~(2) (2015) 295--307.

\bibitem{sree2021novel}
V.~Sree, J.~Mapes, S.~Dua, O.~S. Lih, J.~E. Koh, E.~J. Ciaccio, U.~R. Acharya,
  et~al., A novel machine learning framework for automated detection of
  arrhythmias in ecg segments, Journal of Ambient Intelligence and Humanized
  Computing (2021) 1--18.

\bibitem{jaini2021tool}
S.~N.~B. Jaini, D.~Lee, S.~Lee, M.~Kim, Y.~Kwon, Tool monitoring of end milling
  based on gap sensor and machine learning, Journal of Ambient Intelligence and
  Humanized Computing (2021) 1--13.

\bibitem{mitchell1996investigation}
R.~S. Mitchell, R.~A. Sherlock, L.~A. Smith, An investigation into the use of
  machine learning for determining oestrus in cows, Computers and electronics
  in agriculture 15~(3) (1996) 195--213.

\bibitem{drewek2021survey}
A.~Drewek-Ossowicka, M.~Pietro{\l}aj, J.~Rumi{\'n}ski, A survey of neural
  networks usage for intrusion detection systems, Journal of Ambient
  Intelligence and Humanized Computing 12~(1) (2021) 497--514.

\bibitem{sajja2021image}
T.~K. Sajja, H.~K. Kalluri, Image classification using regularized
  convolutional neural network design with dimensionality reduction modules:
  Rcnn--drm, Journal of Ambient Intelligence and Humanized Computing (2021)
  1--12.

\bibitem{fakhfakh2020prognet}
M.~Fakhfakh, B.~Bouaziz, F.~Gargouri, L.~Chaari, Prognet: Covid-19 prognosis
  using recurrent and convolutional neural networks, The Open Medical Imaging
  Journal 12~(1) (2020).

\bibitem{li2018deep}
Y.~Li, H.~Zhang, X.~Xue, Y.~Jiang, Q.~Shen, Deep learning for remote sensing
  image classification: A survey, Wiley Interdisciplinary Reviews: Data Mining
  and Knowledge Discovery 8~(6) (2018) e1264.

\bibitem{ji20123d}
S.~Ji, W.~Xu, M.~Yang, K.~Yu, 3d convolutional neural networks for human action
  recognition, IEEE transactions on pattern analysis and machine intelligence
  35~(1) (2012) 221--231.

\bibitem{hara2015analysis}
K.~Hara, D.~Saito, H.~Shouno, Analysis of function of rectified linear unit
  used in deep learning, in: 2015 international joint conference on neural
  networks (IJCNN), IEEE, 2015, pp. 1--8.

\bibitem{lau2018review}
M.~M. Lau, K.~H. Lim, Review of adaptive activation function in deep neural
  network, in: 2018 IEEE-EMBS Conference on Biomedical Engineering and Sciences
  (IECBES), IEEE, 2018, pp. 686--690.

\bibitem{apicella2021survey}
A.~Apicella, F.~Donnarumma, F.~Isgr{\`o}, R.~Prevete, A survey on modern
  trainable activation functions, Neural Networks 138 (2021) 14--32.

\bibitem{butts2003network}
C.~T. Butts, Network inference, error, and informant (in) accuracy: a bayesian
  approach, social networks 25~(2) (2003) 103--140.

\bibitem{andrieu2010particle}
C.~Andrieu, A.~Doucet, R.~Holenstein, Particle markov chain monte carlo
  methods, Journal of the Royal Statistical Society: Series B (Statistical
  Methodology) 72~(3) (2010) 269--342.

\bibitem{robert2013monte}
C.~Robert, G.~Casella, Monte Carlo statistical methods, Springer Science \&
  Business Media, 2013.

\bibitem{chaari14}
L.~Chaari, H.~Batatia, N.~Dobigeon, J.-Y. Tourneret, A hierarchical
  sparsity-smoothness bayesian model for l0+l1+l2 regularization, in: IEEE
  International Conference on Acoustics, Speech and Signal Processing (ICASSP),
  2014, pp. 1901--1905.

\bibitem{chaari19}
L.~Chaari, A bayesian grouplet transform, Signal, Image and Video Processing 13
  (2019) 871–878.

\bibitem{fakhfakh2022efficient}
M.~Fakhfakh, B.~Bouaziz, L.~Chaari, F.~Gargouri, Efficient bayesian learning of
  sparse deep artificial neural networks, in: International Symposium on
  Intelligent Data Analysis, Springer, 2022, pp. 78--88.

\bibitem{fakhfakh2022bayesian}
M.~Fakhfakh, B.~Bouaziz, F.~Gargouri, L.~Chaari, Bayesian optimization using
  hamiltonian dynamics for sparse artificial neural networks, in: 2022 IEEE
  19th International Symposium on Biomedical Imaging (ISBI), IEEE, 2022, pp.
  1--4.

\bibitem{fakhfakh2022nonsmooth}
M.~Fakhfakh, L.~Chaari, B.~Bouaziz, F.~Gargouri, Non-smooth bayesian learning
  for artificial neural networks, Journal of Ambient Intelligence and Humanized
  Computing (2022).

\bibitem{university1988continuous}
U.~of~Illinois at Urbana-Champaign. Center~for Supercomputing~Research,
  Development, G.~Cybenko, Continuous valued neural networks with two hidden
  layers are sufficient, 1988.

\bibitem{xiao2005simple}
F.~Xiao, Y.~Honma, T.~Kono, A simple algebraic interface capturing scheme using
  hyperbolic tangent function, International journal for numerical methods in
  fluids 48~(9) (2005) 1023--1040.

\bibitem{cybenko1989approximation}
G.~Cybenko, Approximation by superpositions of a sigmoidal function,
  Mathematics of control, signals and systems 2~(4) (1989) 303--314.

\bibitem{ding2018activation}
B.~Ding, H.~Qian, J.~Zhou, Activation functions and their characteristics in
  deep neural networks, in: 2018 Chinese control and decision conference
  (CCDC), IEEE, 2018, pp. 1836--1841.

\bibitem{harrington1993sigmoid}
P.~d.~B. Harrington, Sigmoid transfer functions in backpropagation neural
  networks, Analytical Chemistry 65~(15) (1993) 2167--2168.

\bibitem{chadha2002fractional}
S.~Chadha, Fractional programming with absolute-value functions, European
  Journal of Operational Research 141~(1) (2002) 233--238.

\bibitem{bengio1994learning}
Y.~Bengio, P.~Simard, P.~Frasconi, Learning long-term dependencies with
  gradient descent is difficult, IEEE transactions on neural networks 5~(2)
  (1994) 157--166.

\bibitem{glorot2011deep}
X.~Glorot, A.~Bordes, Y.~Bengio, Deep sparse rectifier neural networks, in:
  Proceedings of the fourteenth international conference on artificial
  intelligence and statistics, JMLR Workshop and Conference Proceedings, 2011,
  pp. 315--323.

\bibitem{gulcehre2016noisy}
C.~Gulcehre, M.~Moczulski, M.~Denil, Y.~Bengio, Noisy activation functions, in:
  International conference on machine learning, PMLR, 2016, pp. 3059--3068.

\bibitem{nie2011multistability}
X.~Nie, J.~Cao, Multistability of second-order competitive neural networks with
  nondecreasing saturated activation functions, IEEE Transactions on Neural
  Networks 22~(11) (2011) 1694--1708.

\bibitem{montalto2015linear}
A.~Montalto, G.~Tessitore, R.~Prevete, A linear approach for sparse coding by a
  two-layer neural network, Neurocomputing 149 (2015) 1315--1323.

\bibitem{maas2013rectifier}
A.~L. Maas, A.~Y. Hannun, A.~Y. Ng, et~al., Rectifier nonlinearities improve
  neural network acoustic models, in: Proc. icml, Vol.~30, Atlanta, Georgia,
  USA, 2013, p.~3.

\bibitem{konda2014zero}
K.~Konda, R.~Memisevic, D.~Krueger, Zero-bias autoencoders and the benefits of
  co-adapting features, arXiv preprint arXiv:1402.3337 (2014).

\bibitem{dugas2000incorporating}
C.~Dugas, Y.~Bengio, F.~B{\'e}lisle, C.~Nadeau, R.~Garcia, Incorporating
  second-order functional knowledge for better option pricing, Advances in
  neural information processing systems 13 (2000).

\bibitem{Shah2016deep}
A.~Shah, E.~Kadam, H.~Shah, S.~Shinde, S.~Shingade, Deep residual networks with
  exponential linear unit, in: Proceedings of the third international symposium
  on computer vision and the internet, 2016, pp. 59--65.

\bibitem{ccelebi2020new}
M.~{\c{C}}ELEB{\.I}, M.~CEYLAN, New complex valued activationfunctions: Complex
  modifiedswish, complex e-swish andcomplex flatten-tswish., International
  Journal of Advanced Research in Computer Science 11~(2) (2020).

\bibitem{chieng2018flatten}
H.~H. Chieng, N.~Wahid, O.~Pauline, S.~R.~K. Perla, Flatten-t swish: a
  thresholded relu-swish-like activation function for deep learning,
  International Journal of Advances in Intelligent Informatics 4~(2) (2018)
  76--86.

\bibitem{chen1996feedforward}
C.-T. Chen, W.-D. Chang, A feedforward neural network with function shape
  autotuning, Neural networks 9~(4) (1996) 627--641.

\bibitem{guarnieri1995multilayer}
S.~Guarnieri, Multilayer neural networks with adaptive spline-based activation
  functions, in: Proceedings of Word Congress on Neural Networks WCNN'95,
  Washington, DC, July, 1995, pp. 17--21.

\bibitem{piazza1992artificial}
F.~Piazza, A.~Uncini, M.~Zenobi, Artificial neural networks with adaptive
  polynomial activation function (1992).

\bibitem{trottier2017parametric}
L.~Trottier, P.~Giguere, B.~Chaib-Draa, Parametric exponential linear unit for
  deep convolutional neural networks, in: 2017 16th IEEE International
  Conference on Machine Learning and Applications (ICMLA), IEEE, 2017, pp.
  207--214.

\bibitem{qiu2018fre}
S.~Qiu, X.~Xu, B.~Cai, Frelu: flexible rectified linear units for improving
  convolutional neural networks, in: 2018 24th international conference on
  pattern recognition (icpr), IEEE, 2018, pp. 1223--1228.

\bibitem{he2015delving}
K.~He, X.~Zhang, S.~Ren, J.~Sun, Delving deep into rectifiers: Surpassing
  human-level performance on imagenet classification, in: Proceedings of the
  IEEE international conference on computer vision, 2015, pp. 1026--1034.

\bibitem{nader2020searching}
A.~Nader, D.~Azar, Searching for activation functions using a self-adaptive
  evolutionary algorithm, in: Proceedings of the 2020 Genetic and Evolutionary
  Computation Conference Companion, 2020, pp. 145--146.

\bibitem{basirat2018quest}
M.~Basirat, P.~M. Roth, The quest for the golden activation function,
  Proceedings of the ARW \& OAGM Workshop (2019) 1--16.

\bibitem{maguolo2021ense}
G.~Maguolo, L.~Nanni, S.~Ghidoni, Ensemble of convolutional neural networks
  trained with different activation functions, Expert Systems with Applications
  166 (2021) 114048.

\bibitem{sutfeld2020adapt}
L.~R. Sutfeld, F.~Brieger, H.~Finger, S.~Fullhase, G.~Pipa, Adaptive blending
  units: Trainable activation functions for deep neural networks, in: Science
  and Information Conference, Springer, 2020, pp. 37--50.

\bibitem{scardapane2019kafnets}
S.~Scardapane, S.~Van~Vaerenbergh, S.~Totaro, A.~Uncini, Kafnets: Kernel-based
  non-parametric activation functions for neural networks, Neural Networks 110
  (2019) 19--32.

\bibitem{scardapane2018complex}
S.~Scardapane, S.~Van~Vaerenbergh, A.~Hussain, A.~Uncini, Complex-valued neural
  networks with nonparametric activation functions, IEEE Transactions on
  Emerging Topics in Computational Intelligence 4~(2) (2018) 140--150.

\bibitem{ertuugrul2018novel}
{\"O}.~F. Ertu{\u{g}}rul, A novel type of activation function in artificial
  neural networks: Trained activation function, Neural Networks 99 (2018)
  148--157.

\bibitem{castillo2014review}
O.~Castillo, P.~Melin, A review on interval type-2 fuzzy logic applications in
  intelligent control, Information Sciences 279 (2014) 615--631.

\bibitem{scardapane2017learning}
S.~Scardapane, M.~Scarpiniti, D.~Comminiello, A.~Uncini, Learning activation
  functions from data using cubic spline interpolation, in: Italian Workshop on
  Neural Nets, Springer, 2017, pp. 73--83.

\bibitem{wang2019reltanh}
X.~Wang, Y.~Qin, Y.~Wang, S.~Xiang, H.~Chen, Reltanh: An activation function
  with vanishing gradient resistance for sae-based dnns and its application to
  rotating machinery fault diagnosis, Neurocomputing 363 (2019) 88--98.

\bibitem{maguolo2021ensemble}
G.~Maguolo, L.~Nanni, S.~Ghidoni, Ensemble of convolutional neural networks
  trained with different activation functions, Expert Systems with Applications
  166 (2021) 114048.

\bibitem{fakhfakh2022bayesian1}
M.~Fakhfakh, B.~Bouaziz, H.~Batatia, L.~Chaari, Bayesian optimization for
  sparse artificial neural networks: Application to change detection in remote
  sensing, in: Proceedings of International Conference on Information
  Technology and Applications, Springer, 2022, pp. 39--49.

\bibitem{Chaari_tsp_2016}
L.~Chaari, J.-Y. Tourneret, C.~Chaux, H.~Batatia, A {H}amiltonian {M}onte
  {C}arlo method for non-smooth energy sampling, IEEE Trans. on Signal Process.
  64~(21) (2016) 5585 -- 5594.

\bibitem{afshar2021covid}
P.~Afshar, S.~Heidarian, N.~Enshaei, F.~Naderkhani, M.~J. Rafiee, A.~Oikonomou,
  F.~B. Fard, K.~Samimi, K.~N. Plataniotis, A.~Mohammadi, Covid-ct-md, covid-19
  computed tomography scan dataset applicable in machine learning and deep
  learning, Scientific Data 8~(1) (2021) 121.

\bibitem{kayed2020classification}
M.~Kayed, A.~Anter, H.~Mohamed, Classification of garments from fashion mnist
  dataset using cnn lenet-5 architecture, in: 2020 international conference on
  innovative trends in communication and computer engineering (ITCE), IEEE,
  2020, pp. 238--243.

\bibitem{ccalik2018cifar}
R.~C. {\c{C}}alik, M.~F. Demirci, Cifar-10 image classification with
  convolutional neural networks for embedded systems, in: 2018 IEEE/ACS 15th
  International Conference on Computer Systems and Applications (AICCSA), IEEE,
  2018, pp. 1--2.

\bibitem{klambauer2017self}
G.~Klambauer, T.~Unterthiner, A.~Mayr, S.~Hochreiter, Self-normalizing neural
  networks, Advances in neural information processing systems 30 (2017).

\bibitem{lecun1998gradient}
Y.~LeCun, L.~Bottou, Y.~Bengio, P.~Haffner, Gradient-based learning applied to
  document recognition, Proceedings of the IEEE 86~(11) (1998) 2278--2324.

\bibitem{muhammad2018pre}
U.~Muhammad, W.~Wang, S.~P. Chattha, S.~Ali, Pre-trained vggnet architecture
  for remote-sensing image scene classification, in: 24th International
  Conference on Pattern Recognition (ICPR), 2018, pp. 1622--1627.

\bibitem{ioffe2015batch}
S.~Ioffe, C.~Szegedy, Batch normalization: Accelerating deep network training
  by reducing internal covariate shift, in: International conference on machine
  learning, PMLR, 2015, pp. 448--456.

\bibitem{xu20101}
Z.~Xu, H.~Zhang, Y.~Wang, X.~Chang, Y.~Liang, L 1/2 regularization, Science
  China Information Sciences 53~(6) (2010) 1159--1169.

\bibitem{srivastava2014dropout}
N.~Srivastava, G.~Hinton, A.~Krizhevsky, I.~Sutskever, R.~Salakhutdinov,
  Dropout: a simple way to prevent neural networks from overfitting, The
  journal of machine learning research 15~(1) (2014) 1929--1958.

\end{thebibliography}

\end{document}